\documentclass[10pt,twocolumn,letterpaper]{article}

\usepackage{cvpr}              
\usepackage{amsmath,amssymb,amsfonts,amsthm,mathtools,bbm,bm}
\usepackage{xifthen}
\usepackage{array,multirow}

\newcolumntype{M}[1]{>{\centering\arraybackslash}m{#1}}

\definecolor{cvprblue}{rgb}{0.21,0.49,0.74}
\usepackage[pagebackref,breaklinks,colorlinks,allcolors=cvprblue]{hyperref}

\def\paperID{36} 
\def\confName{CVPR}
\def\confYear{2025}

\title{%
Making Every Event Count: \\ Balancing Data Efficiency and Accuracy in Event Camera  Subsampling}

\author{
Hesam Araghi  \quad
Jan van Gemert \quad
Nergis Tomen\\
{Computer Vision Lab, Delft University of Technology}\\
{\tt\small \{h.araghi, j.c.vangemert, n.tomen\}@tudelft.nl}}

\newcommand{\Ecal}{\mathcal{E}}

\newcommand{\Rbb}{\mathbb{R}}

\newcommand{\ratiospatialh}{{r_x}}
\newcommand{\ratiospatialv}{{r_y}}
\newcommand{\ratiospatialhoffset}{{r_{x,0}}}
\newcommand{\ratiospatialvoffset}{{r_{y,0}}}

\newcommand{\ratiotemporal}{{r_t}}
\newcommand{\temporaloffset}{{\Delta t_0}}
\newcommand{\temporalwindow}{{\textit{w}_t}}
\newcommand{\temporalsubsamplinginterval}{{\Delta t}}

\newcommand{\ratiorandom}{{\rho}}

\newcommand{\temporaldecay}{{\tau}}
\newcommand{\filtersize}{{w_d}}
\newcommand{\threshdensity}{{f^{\textrm{(thresh)}}}}
\newcommand{\randomcoefficient}{{u}}
\newcommand{\densityvalue}{{f}}
\DeclareRobustCommand{\spatialkernel}[1]{%
    \ifthenelse{\isempty{#1}}%
    {s}%
    {{s({#1})}}} 

\newcommand{\eventcountthresh}{{\pol_{EC}^{\mathrm{(thresh)}}}}

\newcommand{\harrisscore}{{h_c}}
\newcommand{\harrisfiltersize}{{w_c}}
\newcommand{\harristhresh}{{h_c^{\mathrm{(thresh)}}}}
\newcommand{\cvharrisblocksize}{{\texttt{blockSize}}}

\newcommand{\cvharrisksize}{{\texttt{ksize}}}

\newcommand{\cvharrisk}{{\texttt{k}}}

\newcommand{\estvoxelgrid}{{V}}
\newcommand{\estbinsnum}{{B}}
\DeclareRobustCommand{\estkernel}[1]{%
    \ifthenelse{\isempty{#1}}%
    {f_\theta^{\mathrm{(EST)}}}%
    {{f_\theta^{\mathrm{(EST)}}({#1})}}} 

\newcommand{\event}{{e}}
\newcommand{\eventset}{{\Ecal}}
\newcommand{\eventit}{i}
\newcommand{\eventnum}[1]{{_{#1}}}
\newcommand{\eventtotal}{{N}}
\newcommand{\hpos}{{x}}
\newcommand{\vpos}{{y}}
\newcommand{\tstamp}{{t}}
\newcommand{\pol}{{p}}
\newcommand{\cameraheight}{{H}}
\newcommand{\camerawidth}{{W}}

\newcommand{\averaged}[1]{{\langle#1\rangle}}

\newcommand{\cornerbasedOMAC}{{%
(2\,\cvharrisksize^2 + 3\, \cvharrisblocksize^2+ 10)\,\harrisfiltersize^2 %
}}

\newcommand{\densitybasedOMAC}{{%
4\,\filtersize^2%
}}

\begin{document}
\maketitle
\begin{abstract}

Event cameras offer high temporal resolution and power efficiency, making them well-suited for edge AI applications. 
However, their high event rates present challenges for data transmission and processing. 
Subsampling methods provide a practical solution, but their effect on downstream visual tasks remains underexplored. 
In this work, we systematically evaluate six hardware-friendly subsampling methods using convolutional neural networks for event video classification on various benchmark datasets. 
We hypothesize that events from high-density regions carry more task-relevant information and are therefore better suited for subsampling. 
To test this, we introduce a simple causal density-based subsampling method, demonstrating improved classification accuracy in sparse regimes. 
Our analysis further highlights key factors affecting subsampling performance, including sensitivity to hyperparameters and failure cases in scenarios with large event count variance. 
These findings provide insights for utilization of hardware-efficient subsampling strategies that balance data efficiency and task accuracy.
The code for this paper will be released at: \href{https://github.com/hesamaraghi/event-camera-subsampling-methods}{https://github.com/hesamaraghi/event-camera-subsampling-methods}.
\end{abstract}    
\section{Introduction}
\label{sec:intro}

Event cameras are visual sensors equipped with pixel arrays capable of capturing changes in brightness intensity with a time resolution in the order of microseconds.
Their power efficiency and high temporal resolution make them ideal candidates for edge AI applications \cite{gallego_event-based_2022,hanover_autonomous_2023}.
However, due to their temporal resolution, event cameras can generate exorbitant events per unit time, placing a heavy load on both the transmission and processing systems.
This high event rate is typically associated with increased power consumption \cite{censi_power-performance_2015}, potentially compromising the camera's power efficiency.

 \begin{figure}[t]
    \centering
    \includegraphics[width=\linewidth]{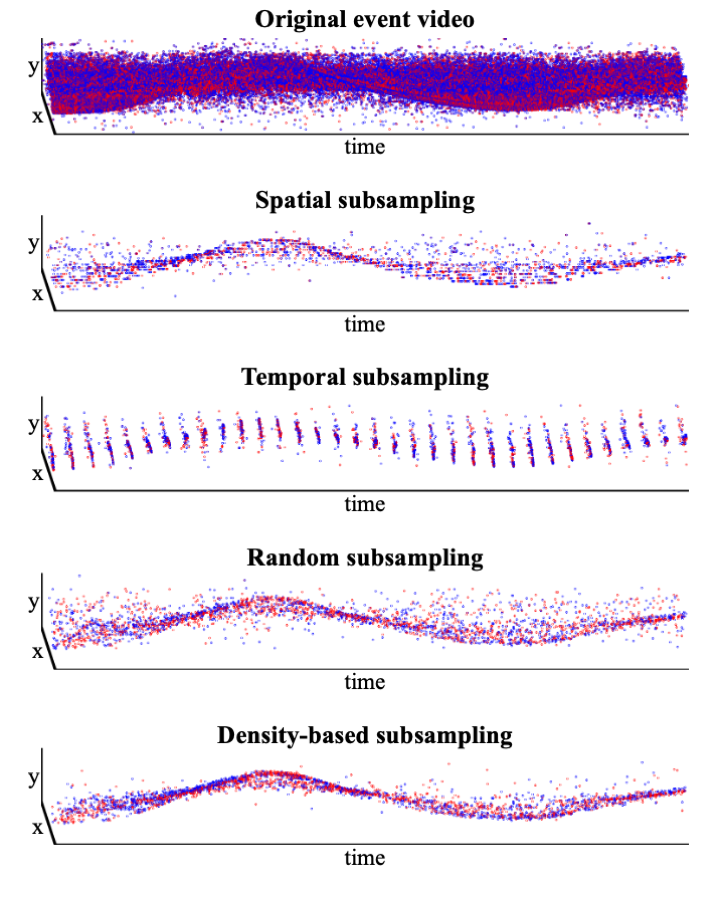}
    \caption{%
Spatial, temporal, random, and density-based subsampling applied to an example video from the DVS-Gesture~\cite{amir_low_2017} dataset, along with the original video. Positive and negative polarity events (red, blue) are plotted in 3 dimensions spanned by pixel coordinates (x, y) and time. Although all subsampled outputs contain the same number of events (3000), their structures vary due to the distinct characteristics of each subsampling method. We evaluate their impact on downstream tasks.
}
    \label{fig:fig_1}
\end{figure}

A simple and effective way to reduce the number of events is through subsampling.
In \cite{araghi_pushing_2024}, the authors have shown that even at higher levels of subsampling, the accuracy of event video classification can remain high.
Nevertheless, there are numerous subsampling methods to consider, some of which are shown in Figure~\ref{fig:fig_1}. In addition, at the hardware level, event cameras may adopt different strategies to manage the event rate.
For example, the event camera introduced in \cite{finateu_a_1280_2020} limits the event rate to a predefined threshold by subsampling events both spatially and temporally. 
Similarly, the event rate controller method in \cite{delbruck_feedback_2021} discards events for a specific time period if the event rate becomes too high.
However, the downstream effects of these subsampling techniques on visual task performance have not been well-studied. 
Choosing the subsampling strategy for a given task remains a challenge, particularly when aiming to minimize subsequent accuracy loss.

This study investigates the effect of different subsampling methods on a downstream task. 
Specifically, we consider object classification as an exemplary downstream task, which plays a key role in many edge AI applications \cite{baldwin_time-ordered_2023,cannici_differentiable_2020,gehrig_end--end_2019,kim_ev-tta_2022,sironi_hats_2018}.
We focus on hardware-friendly subsampling methods which use simple computing operations without demanding high processing power.
In addition, a good subsampling method should reduce the number of events while preserving as much task-relevant information as possible,  thereby minimizing accuracy degradation.
We hypothesize that most relevant information is contained in regions with a high density of events in spatiotemporal space. 
These regions likely correspond to motion in the scene, which typically carries substantial task-relevant information, while isolated events are often triggered by noise.
To test this hypothesis, we propose a density-based subsampling method.
We adopt a causal approach which does not depend on future events, preserving the low latency and requiring no memory-intense event buffering.
The key contributions of this work can be summarized as follows:

\begin{itemize}
    \item We study the accuracy-\#events trade-off for six different subsampling methods on the downstream task of event video classification using convolutional neural networks. 
    Our evaluation is conducted on three benchmark datasets: N-Caltech101, DVS-Gesture, and N-Cars. 
    \item We test the hypothesis that events in high-density regions carry more informative content by proposing a simple causal density-based subsampling method. 
    \item We conduct an in-depth analysis of the factors contributing to performance limitations of subsampling methods.
    Our findings indicate that sensitivity to hyperparameters, such as phase offset, can significantly degrade performance in naive spatial subsampling.
\end{itemize}


\section{Related work}
\label{sec:related_work}

\subsection{Event Rate Reduction}
\label{sec:related work subsec:sparsity}

Reducing the event rate can be achieved through different approaches, including downsampling events by reducing the spatial and/or temporal resolution or filtering out the events.
 In \cite{cohen_spatial_2018}, the authors propose a downsampling method that reduces both temporal and spatial resolution.
 However, their evaluation focuses on classification using a spiking neural network (SNN), which lags behind state-of-the-art methods. 
 This potentially makes their findings less generalizable to other architectures, such as CNNs or transformers. 
In \cite{gu_eventdrop_2021}, the authors use random event subsampling as a data augmentation technique, but they do not study it as a method for subsampling event input. 
The work in \cite{araghi_pushing_2024} investigates the impact of random subsampling on classification accuracy and  the challenges of training CNNs in extremely sparse event regimes. 
In \cite{gruel_event_2022}, authors propose a specific spatial subsampling by normalizing event counts within spatially downscaled regions. 
An event is then triggered when the normalized count exceeds a predefined threshold. 
Other spatial subsampling methods involve using  feedforward SNNs to generate downsampled events \cite{ghosh2024evdownsampling, gruel_performance_2023}. 
However, practical use of these methods would require specialized hardware for SNNs in the camera.


Beyond naive spatial, temporal, and random subsampling, more complex methods take into account interactions between events when deciding which ones to keep. 
Several studies focus on denoising background activity (BA) by filtering out low-density events \cite{guo_low_2023, liu2015design, feng_event_2020, khodamoradi_onon-space_2021}.
However, the  goal is noise removal rather than event subsampling, and the effect of density-based methods on the subsampling is unexplored.

Another relevant line of research involves event-based corner detection methods \cite{vasco_fast_2016, li_fa-harris_2019, mueggler_fast_2017, alzugaray_asynchronous_2018, glover_luvharris_2022}, where most approaches adapt the Harris corner detector \cite{Harris1988} for event streams. 
However, these works primarily focus on corner detection rather than using corners for event subsampling. 
Given the informativeness of corners in standard images \cite{loog_improbability_2010, loog_information_2011}, they could also be leveraged for efficient subsampling of event data.
Despite the widespread use of subsampling methods, such as spatial and temporal subsampling, a systematic comparison of the accuracy performance is lacking.
Thus, we introduce an evaluation study on assessing the effect of different subsampling methods---including density-based and corner-based approaches---across multiple datasets.

\subsection{Event Processing}
\label{sec:related work subsec:processing}

Deep neural networks have significantly advanced event data processing across various applications \cite{zheng_deep_2023}. 
Among them, convolutional neural networks (CNNs) are widely used for event cameras \cite{gallego_event-based_2022, zheng_deep_2023, gehrig_end--end_2019, gehrig_dense_2024}. 
CNNs are particularly attractive due to their computational efficiency compared to more complex models like vision transformers (ViTs) \cite{peng_get_2023, zubic_chaos_2023, klenk_masked_2022} and their improved accuracy over SNNs \cite{vicente-sola_spiking_2023, cohen_spatial_2018, fang2021incorporating} or handcrafted approaches \cite{lagorce_hots_2017, orchard_hfirst_2015, sekikawa_eventnet_2019, sironi_hats_2018}.
Since CNNs require grid-like inputs, event data in address-event representation \cite{mahowald1992vlsi} must first be transformed. 
Various representations exist to do this.
The time surface representation \cite{lagorce_hots_2017} assigns the most recent event timestamp to each pixel, keeping the  temporal information of last events.
Another type of representation is generating frames using counting the events \cite{kogler2009bio,maqueda_event-based_2018}, which aggregates events at each pixel.
This representation is simple and it can keep the spatial information such as edges of the scene, but may discard fine-grained temporal details, potentially causing blurring.
Another approach is voxel grid representation~\cite{zhu_unsupervised_2019}, which divides the time axis into bins and applies predefined kernels to compute bin values.
 A more advanced method, Event Spike Tensor (EST) \cite{gehrig_end--end_2019}, extends this idea by learning a kernel for aggregating the surrounding events into bin values.
The end-to-end learning property of EST allows the model the flexibility of extracting more relevant information from the input event data.
In this paper, we adopt the EST algorithm~\cite{gehrig_end--end_2019} to compare the classification accuracy between the different subsampling methods.

\section{Method}
\label{sec:method}

We represent an event video as a set of events, $\eventset=\{\event\eventnum{\eventit}\}_{\eventit=1}^{\eventtotal}$, where $\eventtotal$ is the total number of events in the video. 
Each event $\event\eventnum{\eventit}$ consists of four values:
$\event\eventnum{\eventit} = (\hpos\eventnum{\eventit},\vpos\eventnum{\eventit},\tstamp\eventnum{\eventit},\pol\eventnum{\eventit})$, where $\hpos\eventnum{\eventit}\in \{1,\ldots,\camerawidth\}$ and $\vpos\eventnum{\eventit} \in \{1,\ldots,\cameraheight\}$ represent the horizontal and vertical spatial positions with $\cameraheight$ and $\camerawidth$ the height and width in pixels, $\tstamp\eventnum{\eventit} \in \mathbb{R}$ is the timestamp, and $\pol\eventnum{\eventit}\in\{-1,1\}$ denotes the polarity of the $\eventit$-th event.
We assert that the time stamps are ordered $\tstamp\eventnum{i}\le\tstamp\eventnum{j}$ for $i\le j$.

\subsection{Event Representation and Training Procedure}
\label{subsec:EST_representation_training}

For object classification in event videos, convolutional neural networks (CNNs) have been well established \cite{baldwin_time-ordered_2023,cannici_differentiable_2020,gehrig_end--end_2019,sironi_hats_2018}
Here, we use CNNs as a proxy for evaluating the information content retained by subsampling methods in the downstream classification task.
We use the EST algorithm~\cite{gehrig_end--end_2019} to represent events in a voxel grid format, making them compatible with convolutional neural networks (CNNs) for classification. 
The voxel grid representation in EST divides the temporal dimension of an event stream into $\estbinsnum$ equally-spaced bins. 
For each bin, events are accumulated into two 2D grid frames---one for each event polarity $\pol\eventnum{\eventit} \in \{-1,1\}$. 
As a result, the final event representation is a voxel grid of size $\estvoxelgrid \in \Rbb^{2\estbinsnum\times\cameraheight\times\camerawidth}$.
The mapping of the events to each voxel in the EST representation is computed by a multilayer perceptron (MLP).
The resulting voxel grid $\estvoxelgrid$ is then fed into a CNN with $2\estbinsnum$ input channels. 
The MLP and the CNN is updated \emph{jointly} during training the model.
This end-to-end training makes the event representation specific to the task, and allows the model to adapt event aggregation based on the input event stream.
We set the number of bins to $\estbinsnum{=}9$, following the original EST paper~\cite{gehrig_end--end_2019}. 
We use ResNet34~\cite{he_deep_2015} as the CNN architecture, initialized with pretrained weights from \texttt{ImageNet-1k\_v1}. 
The input layer is modified to accept $2\estbinsnum{=}18$ channels instead of the original 3. 
The weights of the modified input layer are initialized randomly.
For implementation, we use the original code from the EST paper\footnote{%
\href{https://github.com/uzh-rpg/rpg\_event\_representation\_learning}{https://github.com/uzh-rpg/rpg\_event\_representation\_learning}}.

For each subsampling method, we apply the same subsampling level to both the training and test event videos.
For each video, a subset of events is kept based on the subsampling level and type, which remains fixed throughout the training process.

\subsection{Subsampling Types}
\label{subsec:subsampling-types}

We select subsampling methods which can be easily implemented in hardware. To this end, we prioritize the following characteristics:
\begin{itemize}
    \item The method must be \emph{causal}, meaning it cannot depend on future events.
    This is important for two reasons: 1) A dependence of the subsampling method on future events can severely hamper the latency of the visual task, which is crucial for many real-time, closed-loop or edge AI applications, and 2) Accessing future events would require memory units for buffering, which are known to be high-power components.
    A primary goal of subsampling is to improve power efficiency, and incorporating additional memory units would counteract this objective.
    \item We only allow methods composed of simple computing operations, which are computationally light and do not require significant processing power or memory.
    Using complex algorithms, such as event compression or autoencoders to compute latent representations of the event stream, is not practically feasible for direct hardware implementation.
\end{itemize}

\subsubsection{Spatial Subsampling}
\begin{figure}
    \centering
 \subfloat[spatial]{\includegraphics[width=0.34\linewidth]{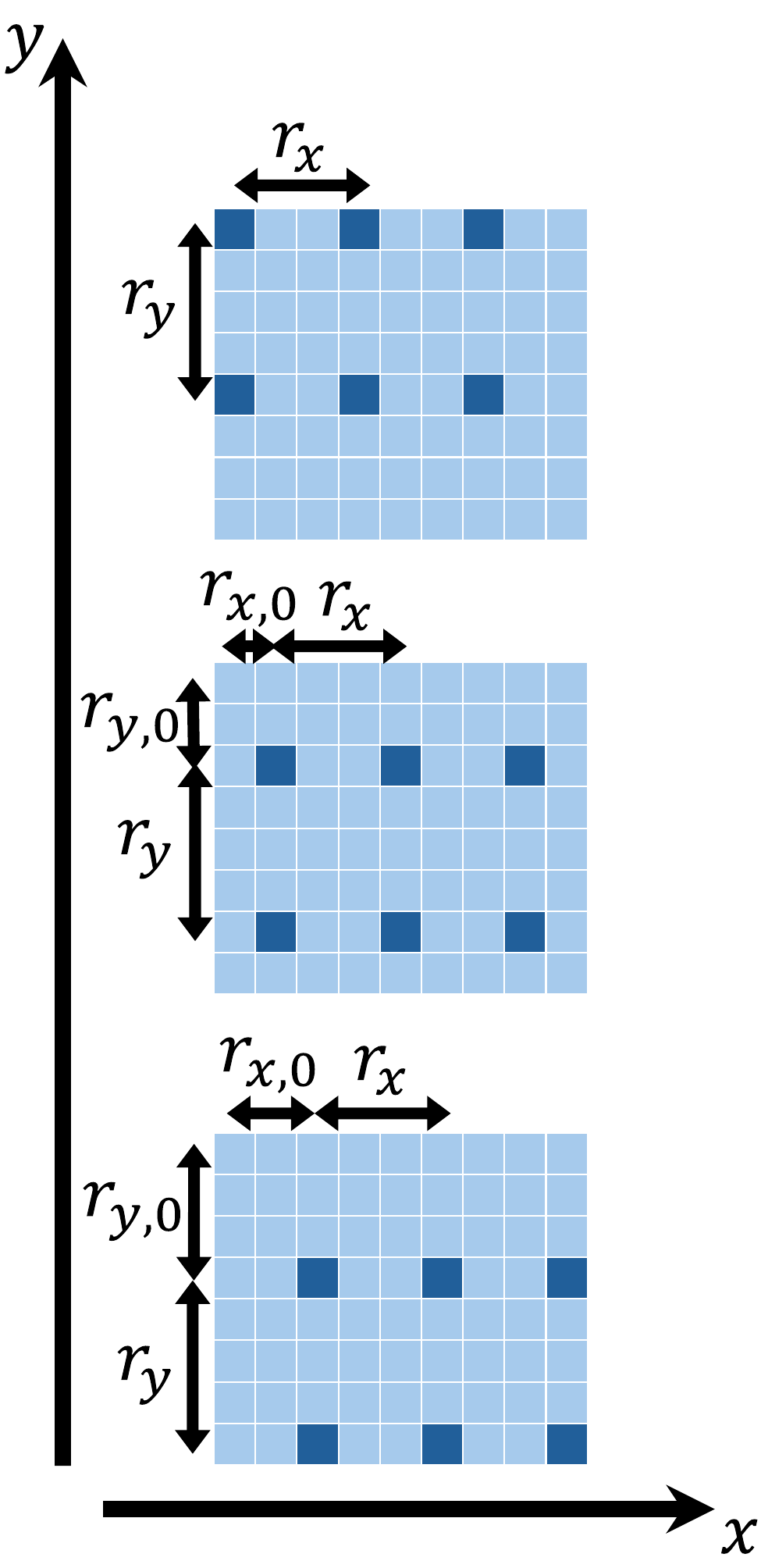}\label{fig:offset:spatial}}
    \hfill
    \subfloat[temporal]{\includegraphics[width=0.65\linewidth]{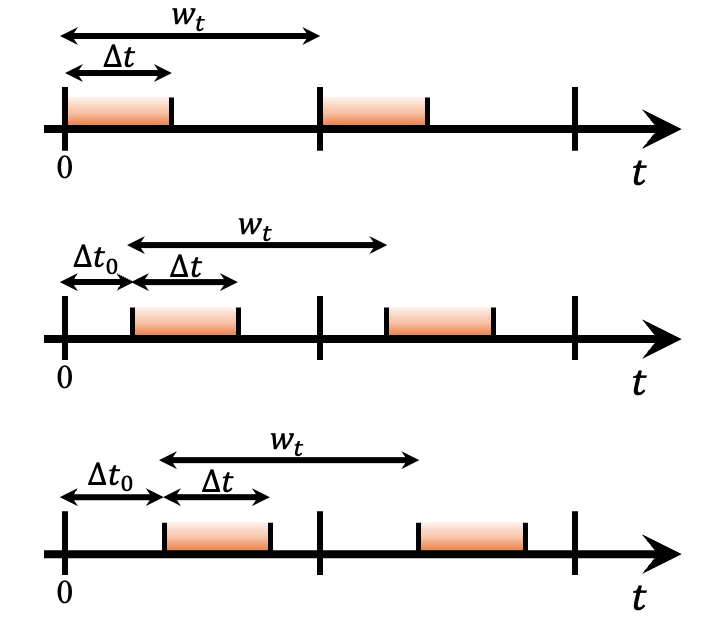}\label{fig:offset:temporal}}
    \caption{\textbf{(a) Spatial subsampling:}
we keep events from every
$\ratiospatialv$-th row vertically and every $\ratiospatialh$-th column horizontally (dark blue pixels). The horizontal and vertical offsets are denoted by $\ratiospatialhoffset$ and $\ratiospatialvoffset$, respectively.  \textbf{(b) Temporal subsampling:} we keep the events within the sampling interval $\temporalsubsamplinginterval$ (colored) in a temporal window of size $\temporalwindow$, where $ \temporaloffset$ is the time offset.
In both cases,
    the topmost subsampling example has zero offset(s).}
\label{fig:offset}
\end{figure}

As illustrated in Fig.~\ref{fig:offset:spatial}, we remove events both horizontally and vertically in spatial subsampling.
Specifically, we retain events from every $\ratiospatialv$-th row vertically and every $\ratiospatialh$-th column horizontally. 
Horizontal and vertical offsets are denoted by $0 \leq \ratiospatialhoffset \leq \ratiospatialh - 1$ and $0 \leq \ratiospatialvoffset \leq \ratiospatialv - 1$, respectively. 
An event $\event=(\hpos,\vpos,\tstamp,\pol)$ is kept if $(\hpos - \ratiospatialhoffset) \mod  \ratiospatialh = 0$, and $(\vpos - \ratiospatialvoffset) \mod \ratiospatialv = 0$, where $\mod$ denotes modulus notation.
The offset values are chosen randomly with equal probability for each training run.

\subsubsection{Temporal Subsampling}
The temporal subsampling is illustrated in Fig.~\ref{fig:offset:temporal}. 
We define a temporal window of size $\temporalwindow$ and divide it by $\ratiotemporal$ to obtain the subsampling interval $\temporalsubsamplinginterval=\frac{\temporalwindow}{\ratiotemporal}$.
We also introduce a time offset, $0 \le \temporaloffset\le\temporalwindow - \temporalsubsamplinginterval$, for subsampling. 
Thus, as shown in Figure~\ref{fig:offset:temporal}, an event $\event=(\hpos,\vpos,\tstamp,\pol)$ is kept if there exists an integer $k$ such that the event's timestamp $\tstamp$ satisfies the following inequality:
\begin{equation} k \temporalwindow + \temporaloffset \le \tstamp < k \temporalwindow + \temporalsubsamplinginterval + \temporaloffset. \label{eq
} \end{equation}
In each training run, the offset $\temporaloffset$ is uniformly drawn from the interval $[0,\temporalwindow - \temporalsubsamplinginterval]$.

\subsubsection{Random Subsampling} 
Each event ${\event=(\hpos,\vpos,\tstamp,\pol)}$ is independently retained with probability $0 \le \ratiorandom \le 1$. 
In each experiment, we first apply random subsampling to the events with probability $\ratiorandom$ and then use the subsampled events consistently throughout training.

\subsubsection{Causal Density-based Subsampling} 
We introduce a  density-based subsampling method with the constraints of being causal, memory-efficient, and computationally inexpensive.

First, we compute a density value $\densityvalue\eventnum{\eventit}^{(\pol\eventnum{\eventit})}$ for each incoming event $\event\eventnum{\eventit} = (\hpos\eventnum{\eventit},\vpos\eventnum{\eventit},\tstamp\eventnum{\eventit},\pol\eventnum{\eventit})$, separately for each polarity $\pol\eventnum{\eventit}$, using the following causal spatiotemporal filtering:
\begin{equation}
\!\!\densityvalue\eventnum{\eventit}^{(\pol\eventnum{\eventit})} = \!\sum_{\substack{j=1 | \pol\eventnum{j} = \pol\eventnum{\eventit}}}^{\eventit} \!\!\spatialkernel{\hpos\eventnum{\eventit} - \hpos\eventnum{j}, \vpos\eventnum{\eventit} - \vpos\eventnum{j}}\exp\left(\frac{\tstamp\eventnum{\eventit}-\tstamp\eventnum{j}}{\temporaldecay}\right),
    \label{eq:spatiotemporal_filtering}
\end{equation}
where $\spatialkernel{\cdot,\cdot}$ is a spatial kernel with filter size $\filtersize \times \filtersize$, while temporal filtering is applied using an exponential kernel with decay parameter $\temporaldecay$.
The formulation in \eqref{eq:spatiotemporal_filtering} allows for separate computation of of spatial filtering $\spatialkernel{\cdot,\cdot}$ and   temporal filtering.
Moreover, the exponential temporal filtering enables recursive computation of the density value $\densityvalue\eventnum{\eventit}^{(\pol\eventnum{\eventit})}$ using the previous value of  $\densityvalue\eventnum{\eventit-1}^{(\pol\eventnum{\eventit-1})}$, eliminating the need for buffering past events.
For  the spatial kernel $\spatialkernel{\cdot,\cdot}$, we use a two-dimensional Gaussian kernel with a standard deviation of $\filtersize / 5$. 
The spatial filter size is set to $\filtersize = 7$, and the temporal decay is set to $\temporaldecay = 30\;\textrm{milliseconds}$.
These values were selected based on the scene dynamics and the camera resolution in the datasets.

\begin{figure}
    \centering
    \subfloat[Original]{\fbox{\includegraphics[width=0.3\linewidth]{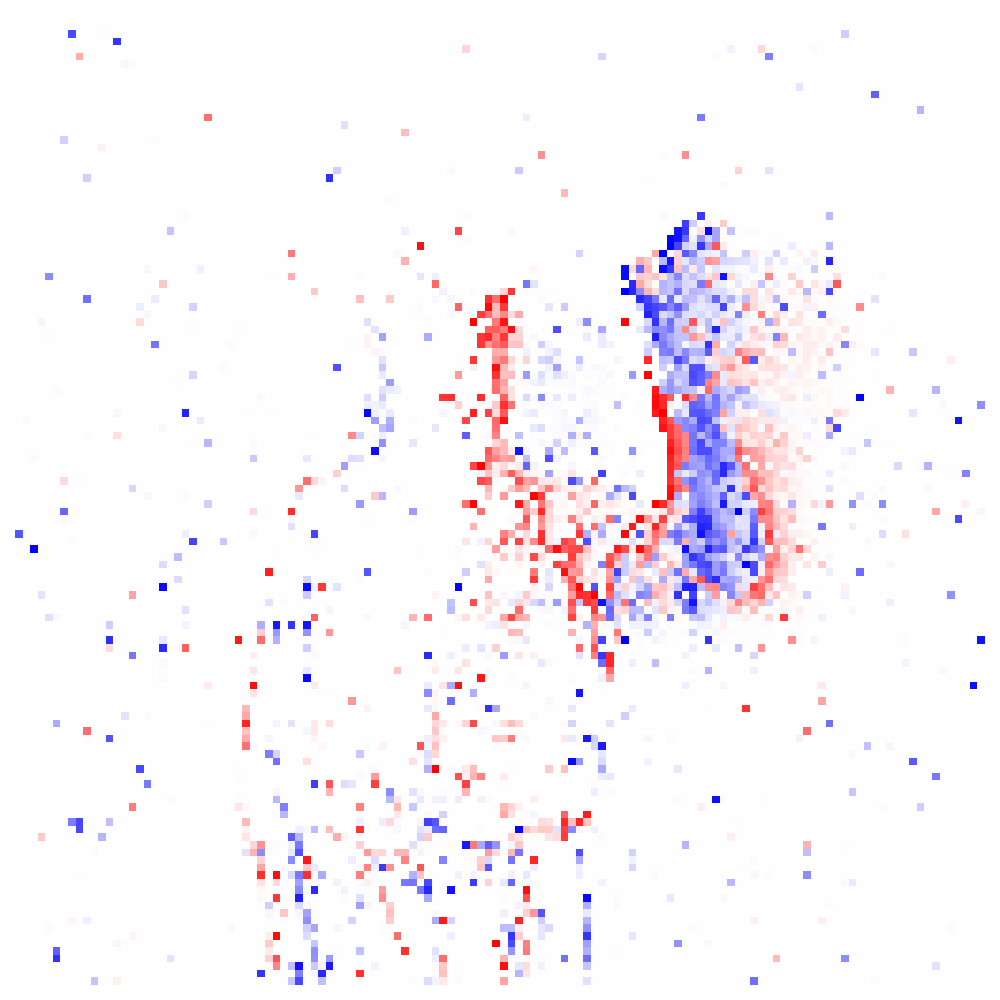}}}
    \hfill        \subfloat[Random threshold]{\fbox{\includegraphics[width=0.3\linewidth]{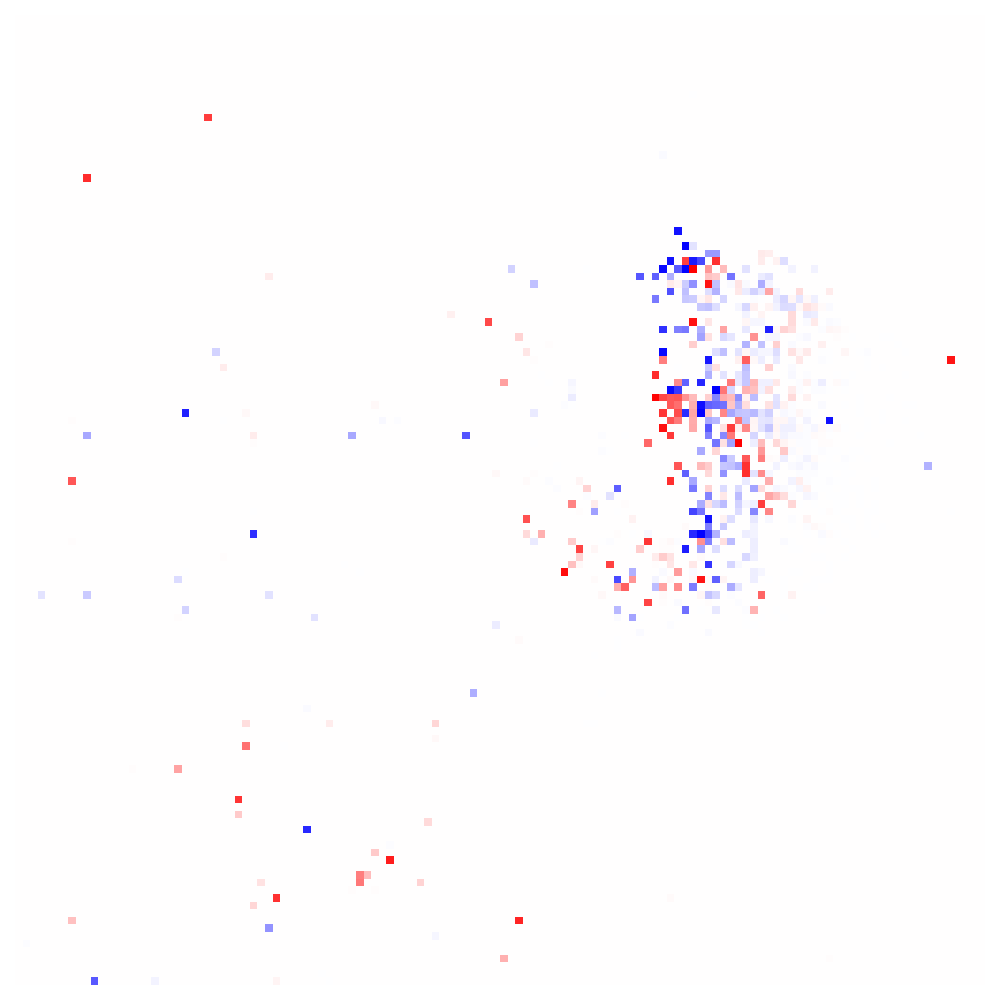}}}
    \hfill
    \subfloat[Fixed threshold]{\fbox{\includegraphics[width=0.3\linewidth]{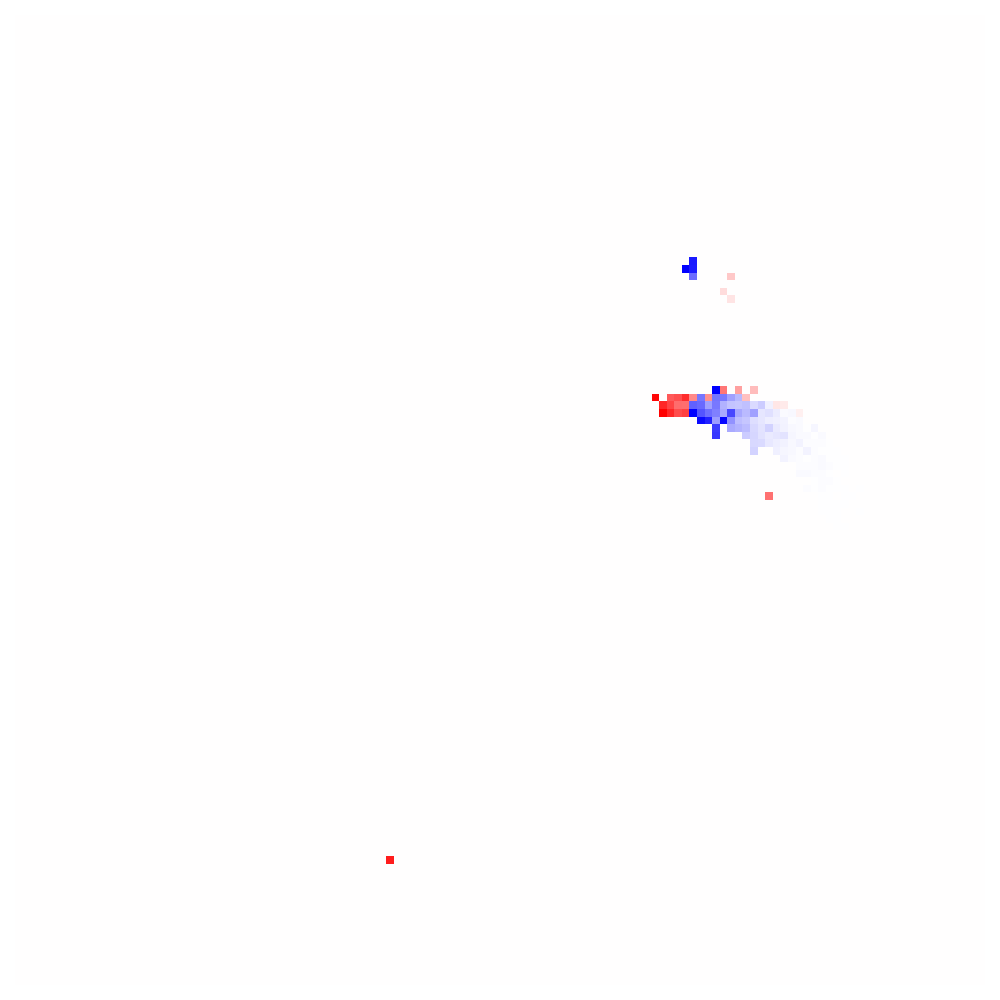}}}
    \caption{Causal density-based subsampling using fixed and random thresholding, retaining a similar number of events. \textbf{(a)} shows the original unfiltered events. 
    Random thresholding \textbf{(b)} preserves the overall shape of the arm and hand movement, while fixed thresholding \textbf{(c)} focuses greedily on a small region near the hand. }
    \label{fig:hard_vs_random_fixed_thresholding}
\end{figure}

A higher density value, $\densityvalue\eventnum{\eventit}^{(\pol\eventnum{\eventit})}$,
shows that an event comes from a denser region.
After computing the density value for the incoming event $\event\eventnum{\eventit} = (\hpos\eventnum{\eventit},\vpos\eventnum{\eventit},\tstamp\eventnum{\eventit},\pol\eventnum{\eventit})$,  we determine whether to keep the event based on a predefined threshold $\threshdensity$.
A higher threshold value results in filtering out more events.
We observe that filtering using a fixed threshold $\threshdensity$ for all events can cause greedy selection from very dense regions only. 
To prevent this, we apply a random thresholding approach.
Instead of using a fixed cutoff, we introduce a random coefficient $\randomcoefficient$, uniformly distributed between 0 and 1, i.e. 
$\randomcoefficient \sim \mathcal{U}(0,1)$.
Particularly, we keep an event $\event\eventnum{\eventit}$ if:
\begin{equation}
\densityvalue\eventnum{\eventit}^{(\pol\eventnum{\eventit})} \ge \randomcoefficient\;\threshdensity.
    \label{eq:density-based:random thresholding}
\end{equation}
Figure~\ref{fig:hard_vs_random_fixed_thresholding} compares the effects of fixed and random thresholding on the representation of subsampled events from a video in the DVS-Gesture dataset. 
As seen in the figure, fixed thresholding results in events being predominantly selected from the densest areas, such as around the hand, while random thresholding allows for a more diverse selection, capturing movement from the entire arm. 

\subsubsection{Event Count Subsampling}
As a baseline method, we consider the Event Count subsampling approach proposed in \cite{gruel_event_2022}, which performs subsampling in the spatial domain. In this method, the full-scale spatial image is divided into non-overlapping windows of size
($\ratiospatialh$,$\ratiospatialv$).
For each window in the full-scale image, there is one corresponding pixel in the subsampled output, effectively spatially downscaling the events.
The averaged polarity of all incoming events within a window is computed by summing the event polarities and dividing by the number of pixels in the window,
$\ratiospatialh\times\ratiospatialv$.
If this normalized event count value crosses a predefined threshold 
$\eventcountthresh$, an event is triggered at the corresponding output pixel.
The polarity of the triggered event is determined by whether the threshold is crossed in an increasing or decreasing manner.

\subsubsection{Corner-based subsampling}
Corners in an image are key interest points that carry high information content \cite{loog_improbability_2010, loog_information_2011}.
Therefore, events that correspond to image corners in an event video can be good candidates for subsampling.
To identify the corners, we adopt the two-dimensional Harris corner detector proposed in \cite{glover_luvharris_2022}. 
This method first introduces an efficient computational event representation called the Threshold-Ordinal Surface (TOS), which produces an 8-bit grayscale representation.
The TOS representation can be updated per each incoming event.
Harris corner detection is then performed using the \texttt{cornerHarris} function from the OpenCV library \cite{opencv_library}.
Similarly, we use the TOS representation to create a 2D event representation. 
With each incoming event, the TOS representation is updated, and a patch of size $\harrisfiltersize\times\harrisfiltersize$, centered at the event's spatial location, is extracted. 
This patch is then passed to the \texttt{cornerHarris} function to compute the Harris score $\harrisscore$ at the center of the patch.
An event is retained as a corner if its Harris score exceeds a predefined threshold, i.e., $\harrisscore>\harristhresh$. 
For parameter selection, we follow the settings from \cite{glover_luvharris_2022}. 
The patch size for the TOS representation is set to 
$\harrisfiltersize{=}7$, while the \texttt{cornerHarris} function parameters are kept at their default values: $\cvharrisblocksize{=}2$, 
$\cvharrisksize{=}3$, and 
$\cvharrisk{=}0.04$. 
The Sobel operator is used to compute the horizontal and vertical derivatives.

\subsection{Event Classification Datasets}
\label{sec:method sebsec:datasets}

\noindent\textbf{N-Caltech101 \cite{orchard_converting_2015}.} 
This dataset is generated by displaying static images from the Caltech101~\cite{fei_learning_2004} dataset in front of an ATIS event camera \cite{posch_atis_2010} and moving the camera in three directions to trigger events. 
It consists of 101 classes.
As all videos follow the same predefined motion pattern, the temporal details of the events do not correlate strongly with class information.

\noindent\textbf{DVS-Gesture \cite{amir_low_2017}.}
This dataset consists of various hand and arm gestures in 11 classes, recorded from 29 subjects under three different lighting conditions. 
The gestures were captured using a fixed DVS128 event camera with a stationary background.
This dataset contains real dynamic motion generated by actual hand and arm movements rather than predefined camera motion.

\noindent\textbf{N-Cars \cite{sironi_hats_2018}.}
This dataset was recorded using an ATIS event camera mounted on a car driving through urban environments. 
It contains two classes: cars and background.
Similar to the DVS-Gesture dataset, the events capture real motion dynamics. 
However, the camera is not fixed, leading to large background variations in the dataset.

\section{Experiments}
\label{sec:experiments}

In this section, we examine the effect of subsampling methods introduced in Subsection~\ref{subsec:subsampling-types} on classification accuracy for the datasets \textbf{N-Caltech101} \cite{orchard_converting_2015}, \textbf{DVS-Gesture} \cite{amir_low_2017}, and \textbf{N-Cars} \cite{sironi_hats_2018}. 
Our choice of datasets spans a range of event-based inputs, from  data with little class-related temporal information to recordings with diverse motion dynamics. 
The selection includes both static and moving cameras, as well as varying background conditions.

Classification performance is evaluated across different subsampling levels.
For each subsampling level, the parameters of the subsampling methods are selected to ensure that the average number of events per video remains similar between the methods.
Specific parameter values for each method are provided in the supplementary material.
For all experiments, we use Adam optimizer \cite{kingma_adam_2017} for training.
The learning rate scheduler is `Reduce on Plateau' with reducing factor 2.
The patience parameter is set to 40 for N-Caltech101 and DVS-Gesture, and 50 for N-Cars. 
Each model is trained for 250 epochs.
For the batch size and learning rate, we follow the suggested parameters in \cite{araghi_pushing_2024} for each datasets. 
We used the same hyperparameters across all subsampling levels.

\begin{figure}[t]
    \centering
        \includegraphics[width=\linewidth]{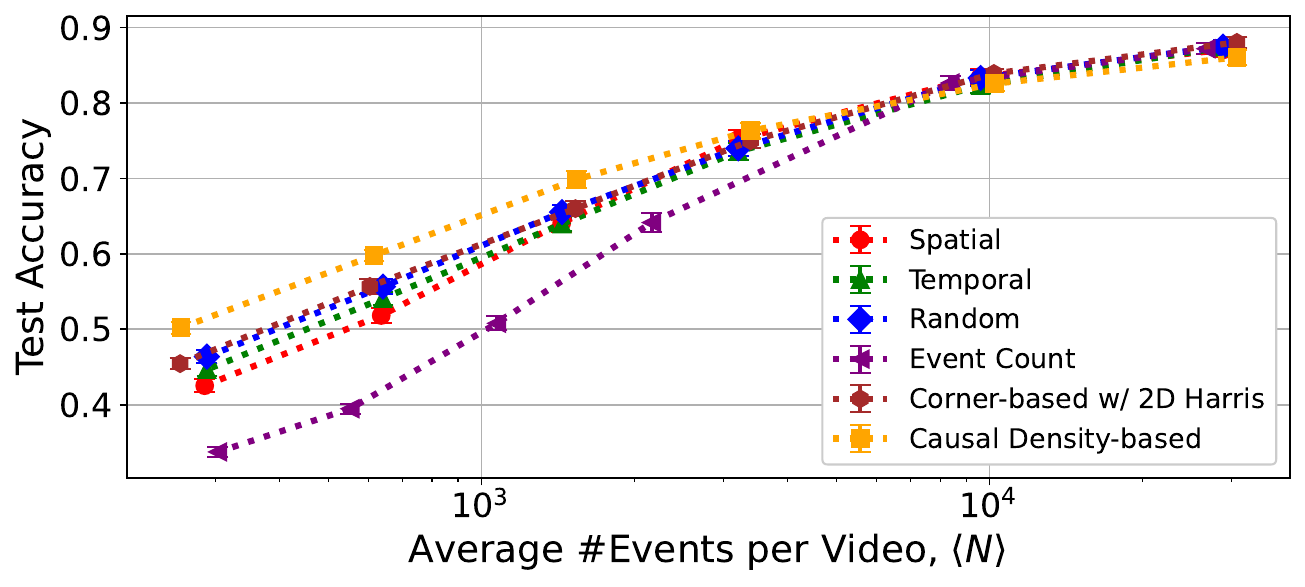}
    \caption{Classification accuracy at different subsampling levels for six subsampling methods on the \textbf{N-Caltech101} dataset. Each curve is the average of 18 independent runs with different seeds. The $x$-axis represents the \textit{average} number of events per video, $\averaged\eventtotal$. Error bars show the standard deviation across runs.}
    \label{fig:sparsity_vs_acc/NCALTECH101}
\end{figure}

\subsection{N-Caltech101 Dataset}
\label{sec:exp:subsec:ncaltech101}

In Figure~\ref{fig:sparsity_vs_acc/NCALTECH101}, we present the classification test accuracy of six subsampling methods for N-Caltech101 dataset.  
The horizontal axis shows the average number of events per video $\averaged\eventtotal$.
The accuracy--\#events curves are generated by evaluating classification test accuracy across different subsampling levels.
Each curve represents the average performance over 18 independent runs with different random seeds.

The results show that input independent methods of spatial, temporal, and random subsampling perform better than the baseline method proposed in \cite{gruel_event_2022}%
\footnote{%
For the implementation of the Event Count method, we used the repository: \href{https://github.com/ameliegruel/EvVisu}{https://github.com/ameliegruel/EvVisu}}.
This highlights the importance of considering simple subsampling techniques as baselines while evaluating more complex approaches.
Interestingly, random subsampling achieves slightly better performance than spatial and temporal subsampling. 
This is noteworthy because most event camera hardware reduces event rates through spatial and/or temporal subsampling, while our results suggest that they may be suboptimal compared to random subsampling.

Among the more complex, input dependent methods---casual density-based and Harris corner-based subsampling---the density-based method achieves the highest accuracy, particularly in the sparser event regimes, while the corner-based method performs comparably to random subsampling. The differences are insightful considering the small magnitude of the standard deviations (error bars).
This result supports the hypothesis that denser regions in event data carry more information, and subsampling from these areas leads to better classification accuracy.


\begin{figure}[t]
    \centering
        \includegraphics[width=\linewidth]{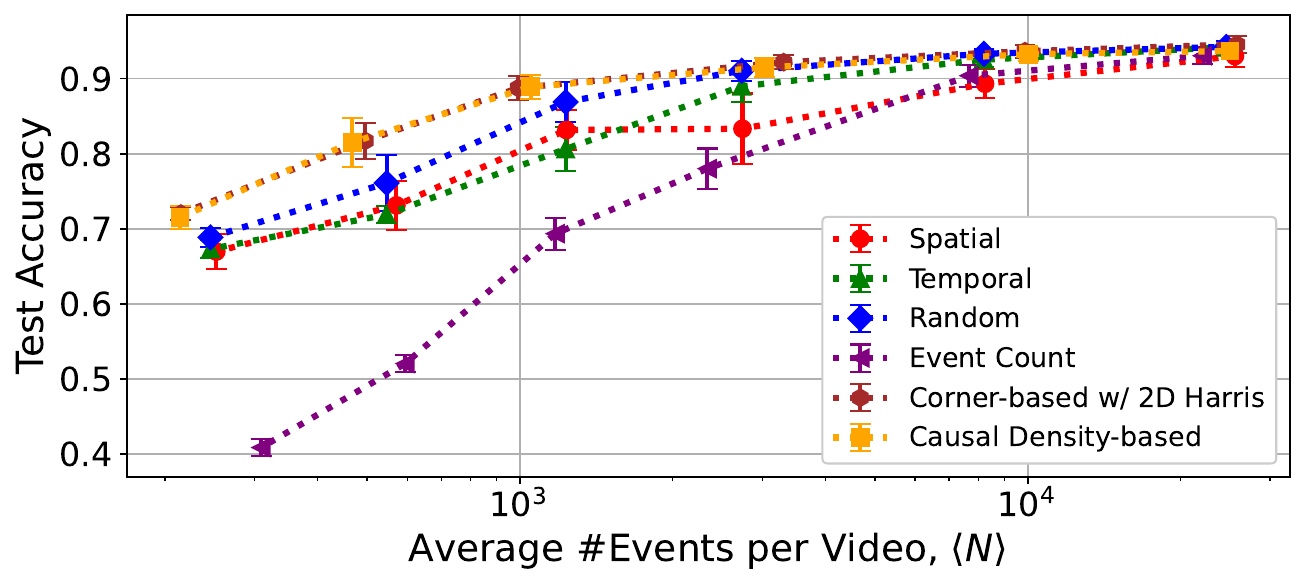}
    \caption{Classification accuracy at different subsampling levels for six subsampling methods on the \textbf{DVS-Gesture} datasets. Each curve is the average of 18 independent runs with different seeds. Error bars show the standard deviation across runs.}
    \label{fig:sparsity_vs_acc/DVSGESTURE}
\end{figure}

\subsection{DVS-Gesture Dataset}
\label{sec:exp:subsec:dvs-gesture}

The accuracy curves for DVS-Gesture dataset is presented in 
Figure~\ref{fig:sparsity_vs_acc/DVSGESTURE} for the six subsampling methods.
The curves are the average of 18 independent runs.
The results for this dataset  align with those of N-Caltech101 in Subsection~\ref{sec:exp:subsec:ncaltech101}.
Specifically, the trivial, input-independent methods---spatial, temporal, and random subsampling---outperform the baseline Event Count method \cite{gruel_event_2022}. Among these, random subsampling has better accuracy than spatial and temporal.
However, in this experiment, the corner-based method achieves similar accuracy to the causal density-based method, with both methods surpassing the other subsampling strategies. 
This suggests that corner events are particularly informative for classifying DVS-Gesture.

\begin{figure}[t]
        \centering
        \includegraphics[width=\linewidth]{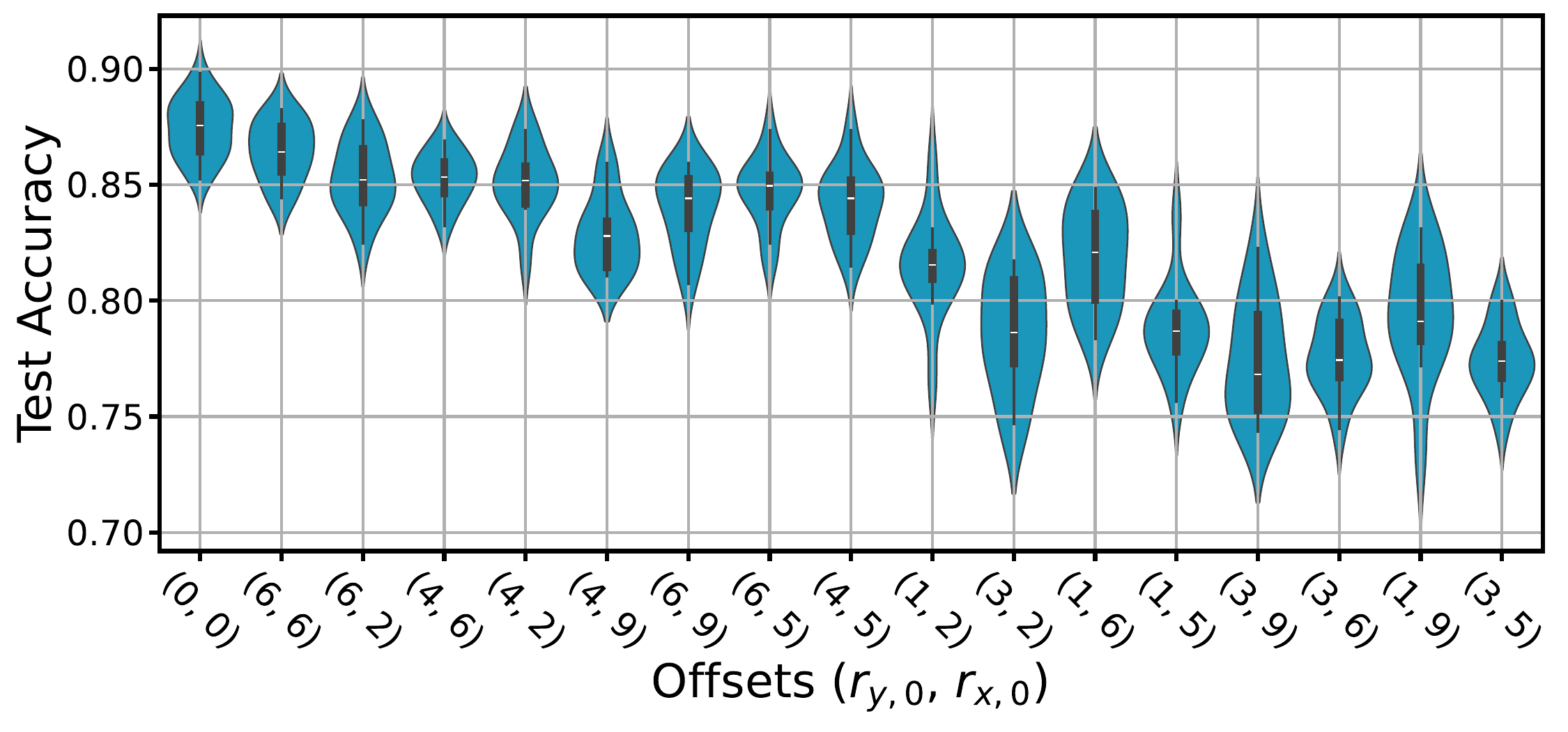}
    \caption{Effect of different sets of horizontal and vertical offsets in spatial subsampling on the test accuracy for DVS-Gesture dataset. 
The horizontal subsampling is   $\ratiospatialh=10$ and vertical is $\ratiospatialv=8$.}
    \label{fig:spatial_offset_8x10}
\end{figure}

An important observation in Figure~\ref{fig:sparsity_vs_acc/DVSGESTURE} is the relatively high standard deviation of accuracies (i.e., diverse performance) for the spatial subsampling method.
By investigating this high variance, we find that it is due to the high sensitivity of spatial subsampling to offset selection.
Figure~\ref{fig:spatial_offset_8x10} illustrates this sensitivity, where we subsample horizontally with $\ratiospatialh=10$ and vertically with $\ratiospatialv=8$.
Test accuracies are computed for different offset configurations, each evaluated with multiple random seeds.
The high performance differences for various offsets shows the high sensitivity of the spatial subsampling to offset selection.
This result suggests that spatial subsampling not only removes high-frequency spatial information but can also distort low-frequency information due to spatial aliasing effects. Combined, our results indicate that naive spatial subsampling at the hardware level may hamper downstream tasks depending on the frequency content of the video.

\begin{figure}[t]
    \centering
    \begin{subfigure}{0.47\textwidth}
        \centering
        \includegraphics[width=\linewidth]{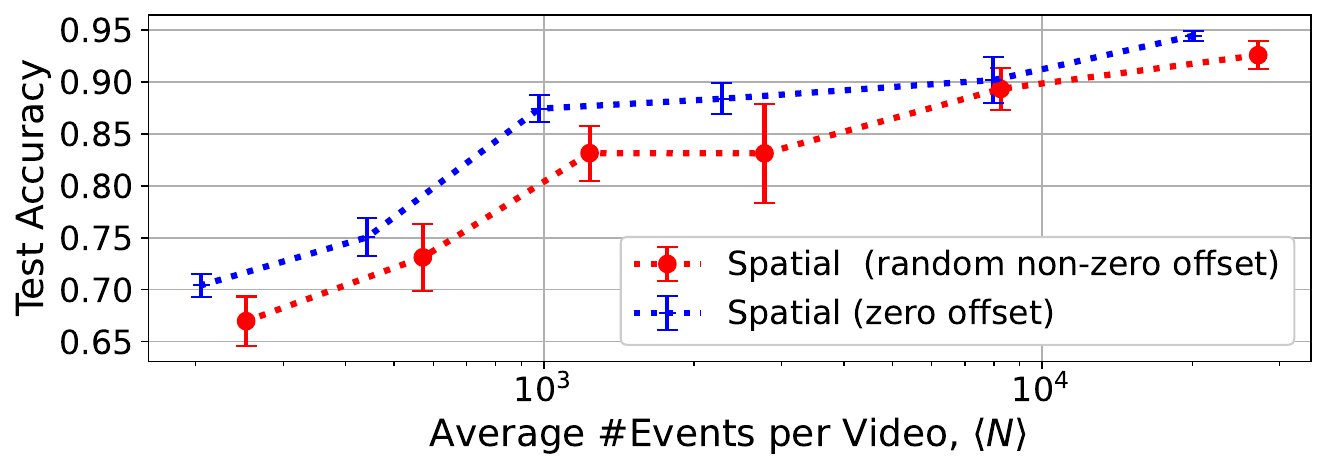}
        \caption{Spatial subsampling}
        \label{fig:offset/spatial_zero_vs_nonzero_offset}
    \end{subfigure}
    \\
    \begin{subfigure}{0.47\textwidth}
        \centering
        \includegraphics[width=\linewidth]{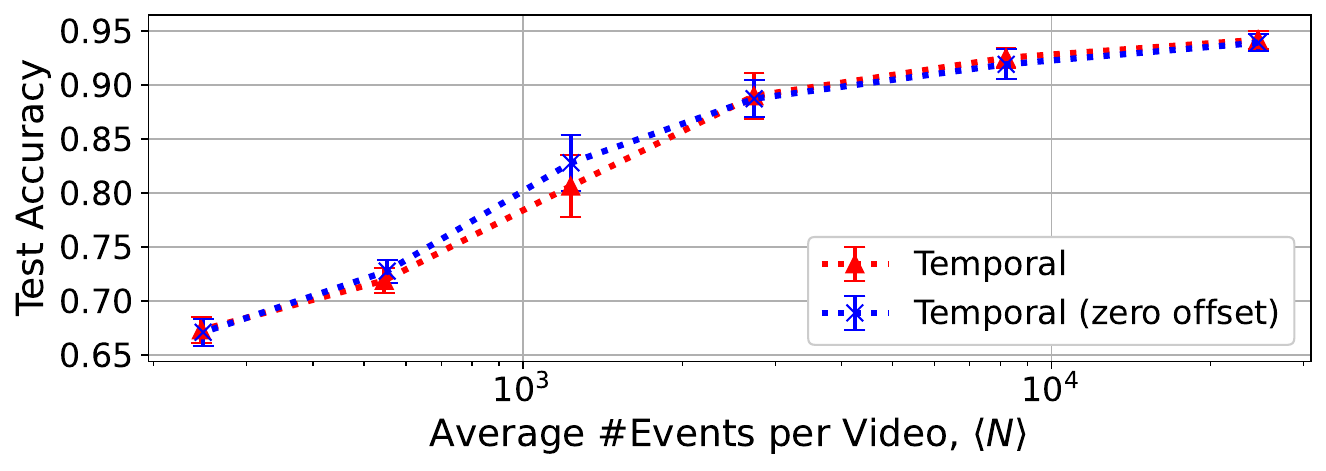}
        \caption{Temporal subsampling}
        \label{fig:offset/temporal_zero_vs_nonzero_offset}
    \end{subfigure}
    \caption{Comparing the offset sensitivity between \textbf{(a)} spatial and \textbf{(b)} temporal subsampling in DVS-Gesture dataset for different levels of sparsity.
    We observe that temporal subsampling is less sensitive to changes in the offset.} \label{fig:offset/spatial_vs_temporal_offset_effect}
\end{figure}

Figure~\ref{fig:offset/spatial_vs_temporal_offset_effect} compares classification accuracy for spatial and temporal subsampling under two conditions: zero offset and random non-zero offset, across different subsampling levels. 
The results show that temporal subsampling has lower sensitivity to offset variations compared to spatial subsampling.
This highlights an advantage of temporal subsampling in terms of offset robustness. 
Given that spatial subsampling is commonly used for event rate reduction---typically by discarding events from specific rows and/or columns---we highlight that temporal subsampling may be a more stable alternative, considering the greater susceptibility of spatial subsampling to the effect of offset.

\begin{figure}[t]
    \centering
    \includegraphics[width=\linewidth]{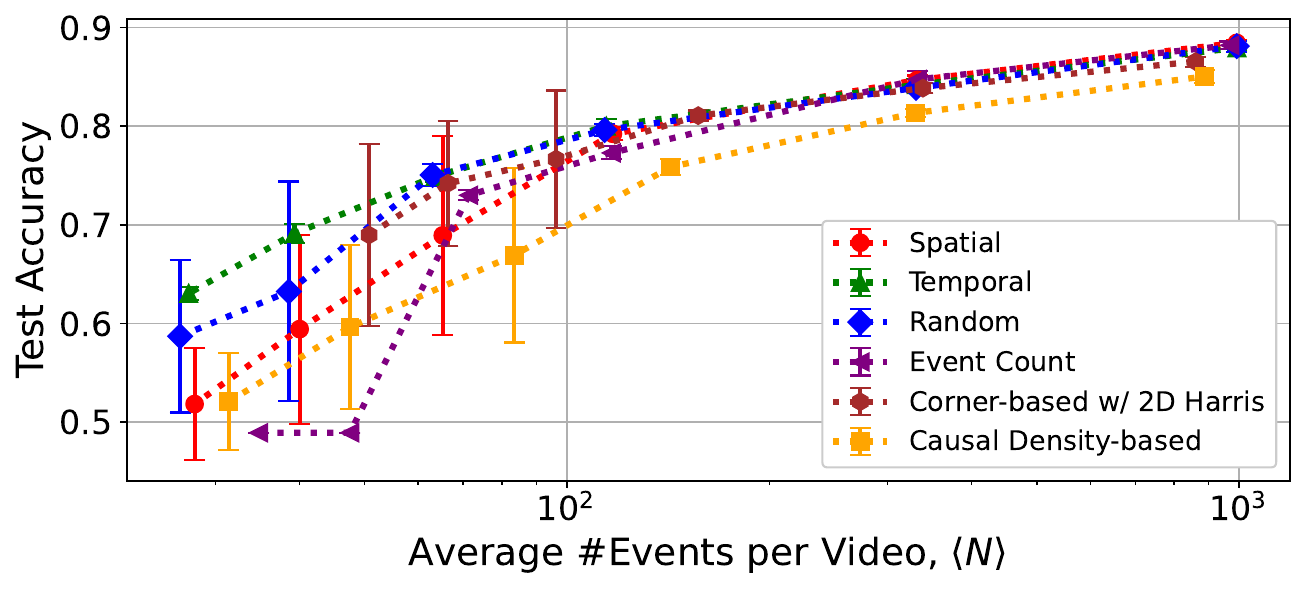}
    \caption{Classification accuracy at different subsampling levels for six subsampling methods on the \textbf{N-Cars} dataset. Each curve is the average of six independent runs with different seeds. Error bars show the standard deviation across runs.}
    \label{fig:low_event_rate/ncars}
\end{figure}

\subsection{N-Cars Dataset}
\label{sec:exp:subsec:n-cars}

In Figure~\ref{fig:low_event_rate/ncars}, we present the results of different subsampling methods on the two-class object classification dataset \mbox{N-Cars}. 
Due to the larger dataset size and higher training time, we average over six independent runs.
The number of events in N-Cars is significantly lower than in N-Caltech101 and DVS-Gesture. 
As a result, we see that subsampling low-rate datasets like N-Cars can lead to performance failure. 
As shown in Figure~\ref{fig:low_event_rate/ncars}, below 100 events per video, most subsampling methods fail to classify the videos. This is both evidenced by the drop in test accuracy towards 50\% (chance level) and the large accuracy variation across runs (error bars). Only temporal subsampling seems to perform slightly better than the others in the sparse regime, and is more stable across runs.

\begin{figure}[t]
    \centering
    \includegraphics[width=\linewidth]{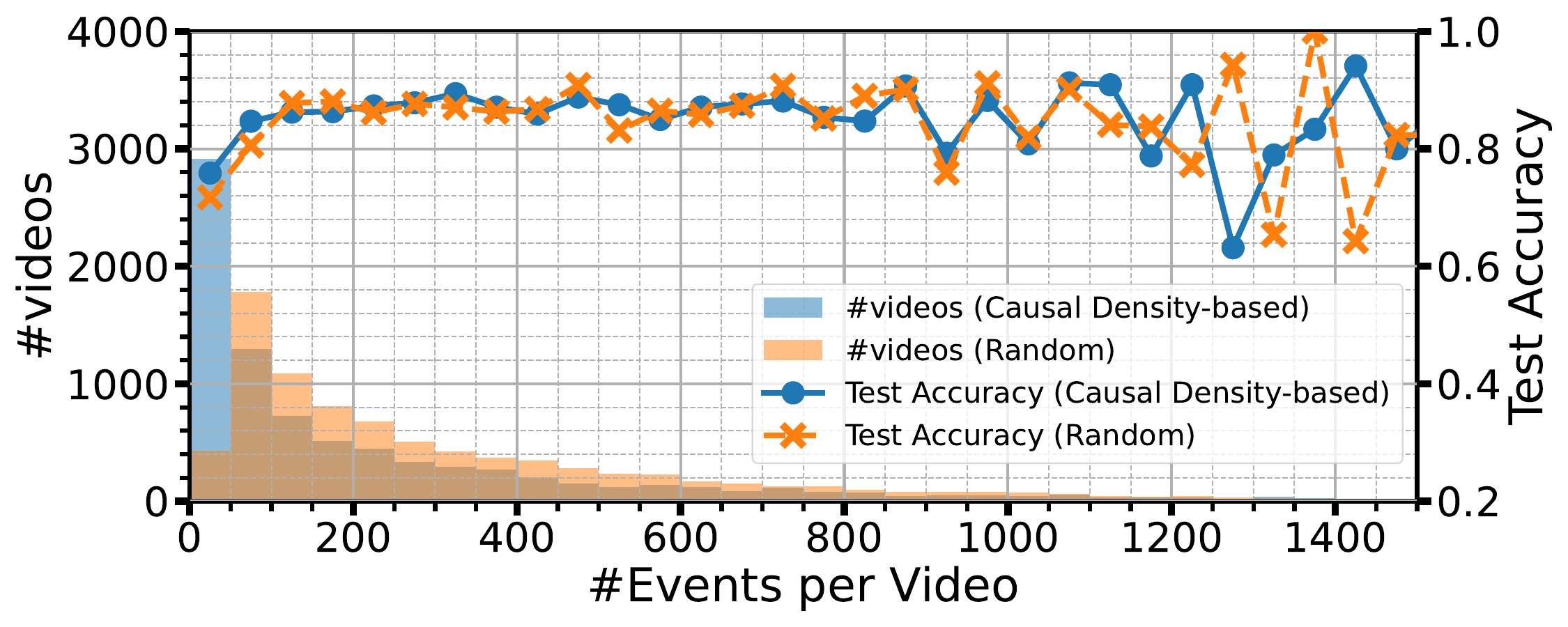}
    \caption{Causal density-based subsampling results in many videos with very low number of events (bar plot, blue) compared to random subsampling (bar plot, orange). Such videos (e.g. first blue bar, with $<50$ events per video, distributed over 18 input channels) achieve lower classification test accuracy (line plots), which in turn lead to lower mean test accuracy for the causal density-based method.}
    \label{fig:low_event_rate/event_histogram_ncars}
    \vspace{-0.3cm}
\end{figure}

Another key observation is the relative accuracy drop in the causal density-based method. This can be attributed to the causal nature of the method, which is agnostic to the total number of events per video.
Since the method cannot know how many events will be recorded before the end of each video, the filter parameters (Eq.~\ref{eq:spatiotemporal_filtering}) as well as the threshold $\threshdensity$ is fixed for the whole dataset.
This means that, in the case of the N-Cars dataset, where the variance of the total number of events across samples is large, the causal density-based subsampling leads to a large number of samples with close to zero events in the video.
This effect is illustrated in Figure~\ref{fig:low_event_rate/event_histogram_ncars}, which shows the histogram (bar plot) of the number of events per video \textit{after} both random and causal density-based subsampling.
Since density values $\densityvalue\eventnum{\eventit}^{(\pol\eventnum{\eventit})}$ in sparse videos are low, more events are removed after applying a fixed threshold to the density values.

\begin{figure}[t]
    \centering
    \includegraphics[width=\linewidth]{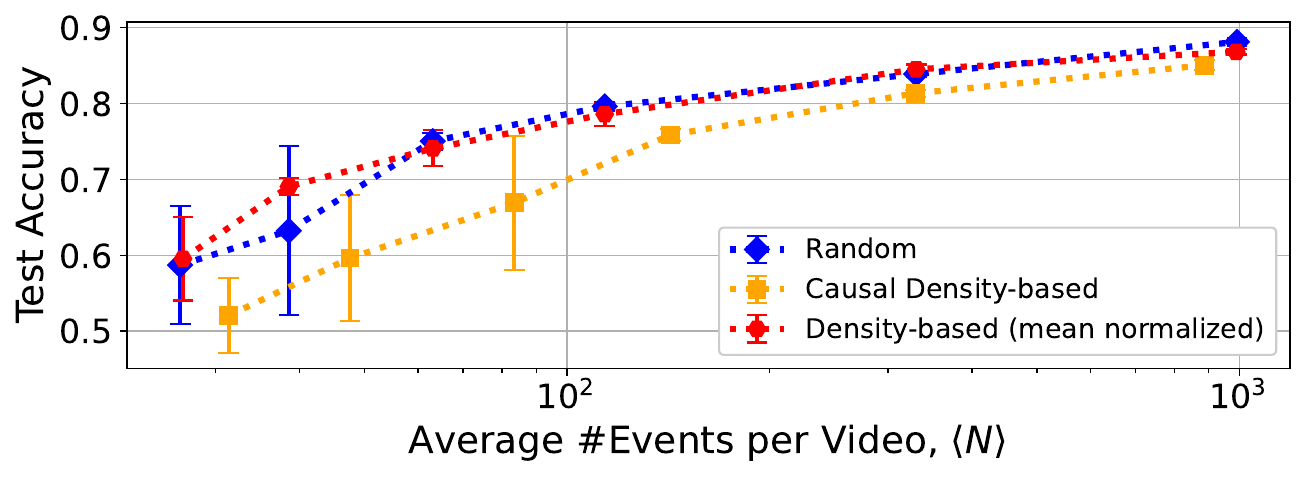}
    \caption{Normalizing the density values $\densityvalue\eventnum{\eventit}^{(\pol\eventnum{\eventit})}$ before thresholding (red) improves the performance of density-based subsampling (yellow). Random subsampling (blue) given for comparison.
    }
    \label{fig:low_event_rate/normalizing_density_based_method}
    \vspace{-0.5cm}
\end{figure}

To mitigate this issue, we normalize the density values before thresholding, preventing excessive subsampling of sparser videos.
As shown in Figure~\ref{fig:low_event_rate/normalizing_density_based_method}, normalizing the density values improves accuracy.
This normalization cost is to lose the causal property of the subsampling process, however, it demonstrates that the density-based subsampling approach can still capture informative events in the N-Cars dataset. 
In practice, a similar improvement can be achieved by simply employing an adaptive threshold, which can, for example, be lowered when the event rate is low and vice versa. 
In this case, the additional memory and power requirements of keeping a running average of the event rate would be implementation-dependent.

\subsection{Performance Overview Across Datasets}
\label{sec:exp:subsec:summary_over_all}

To provide a quantitative comparison between the different subsampling methods, we define a metric evaluating their task performance across all subsampling levels. 
Given that the primary objective of subsampling is to maintain information effectively in settings with a low number of events, we want to place greater emphasis on accuracy in this regime.
To achieve this, we compute the area under the curve (AUC) for the test accuracy as a function of the logarithm of number of events:
\begin{equation}
    \textrm{AUC}_{\textrm{acc-\#events}} = \int \textrm{acc}(\textrm{\#events})d\,(\log_{10}\textrm{\#events}).
    \label{eq:auc_log}
\end{equation}
To normalize this value, we divide the AUC in \eqref{eq:auc_log} by $\textrm{AUC}_{\textrm{acc-\#events}}^{(oracle)}$,  which represents the AUC of an ideal classifier achieving an accuracy of 1.0 across all number of events:
\begin{equation}
    \textrm{nAUC}_{\textrm{acc-\#events}} = \frac{\textrm{AUC}_{\textrm{acc-\#events}}}{ \textrm{AUC}_{\textrm{acc-\#events}}^{(oracle)}}.
    \label{eq:normalized_auc_log}
\end{equation}
The normalized AUC provides a quantitative measure of the overall classification accuracy of subsampling methods, with an increased weight on the low-event regime.
We compute $\textrm{nAUC}_{\textrm{acc-\#events}}$ for each subsampling method using 18 independent runs for N-Caltech101 and DVS-Gesture and 6 for N-Cars.
For N-Cars, we only consider experiments where $\langle \eventtotal \rangle > 50$ to filter out runs with
 highly variant accuracy due to very low total event counts.

Table~\ref{tab:normalized_auc_log} presents the mean and standard deviation of $\textrm{nAUC}_{\textrm{acc-\#events}}$ for different subsampling methods on the three datasets. 
In N-Caltech101, the causal density-based method outperforms other approaches, with random and corner-based subsampling coming in second.
In DVS-Gesture, the causal density-based and corner-based subsampling methods achieve similarly high performance compared to other methods.
For N-Cars, temporal subsampling performs slightly better than other methods, while most approaches achieve similar performance with $\textrm{nAUC} \approx 0.82$. 
The causal density-based method initially performs suboptimally with $\textrm{nAUC} \approx 0.79$. 
However, as analyzed in Subsection~\ref{sec:exp:subsec:n-cars}, its performance improves after normalizing the density values, bringing it back to 0.82.

Taken together, within the class of more advanced subsampling methods we evaluate, corner-based and density-based methods seem to demonstrate the best accuracy-\#event efficiency balance. In terms of computational complexity the causal density-based method displays a slight advantage. Table~\ref{tab:supp:comp_complexities} shows the memory usage and computational cost in multiply–accumulate operations (MACs) for an event video with an spatial resolution of $\cameraheight\times\camerawidth$, and total event number of $\eventtotal$.  
Simple methods such as spatial, temporal, and random subsampling require only $\mathcal{O}(1)$ memory and no specific MAC operation.  
The Event Count method \cite{gruel_event_2022} has higher memory usage compared to these simpler methods.  
For more complex methods, such as corner-based and causal density-based subsampling, memory usage scales with the camera resolution ($\mathcal{O}(\cameraheight\camerawidth)$), while computational cost depends on the square of filter sizes used in these algorithms.  
However, for the same filter size ($\filtersize{=}\harrisfiltersize$), the computational cost of the corner-based method is higher due to the additional operations required for the Harris corner detection algorithm.
More details are provided in supplementary material.

\begin{table}[t]
    \centering
     \caption{Mean and standard deviation  of $\textrm{nAUC}_{\textrm{acc-\#events}}$ for different subsampling methods. The highest value for each dataset is highlighted in {\bf bold}, and the second highest with an \underline{underline}.}
     \label{tab:normalized_auc_log}
      \resizebox{\linewidth}{!}{
        \begin{tabular}{lccc}
        \toprule
        Method & N-Caltech101 & DVS-Gesture & N-Cars \\
        \midrule
        Spatial & 0.697 {\scriptsize $\pm$ 0.005} & 0.827 {\scriptsize $\pm$ 0.016} & 0.818 {\scriptsize $\pm$ 0.009} \\
        Temporal & 0.697 {\scriptsize $\pm$ 0.004} & 0.841 {\scriptsize $\pm$ 0.007} & \textbf{0.827 {\scriptsize $\pm$ 0.002}} \\
        Random & \underline{0.708 {\scriptsize $\pm$ 0.003}} & 0.867 {\scriptsize $\pm$ 0.009} & \underline{0.825 {\scriptsize $\pm$ 0.003}} \\
        Event Count & 0.648 {\scriptsize $\pm$ 0.006} & 0.748 {\scriptsize $\pm$ 0.007} & 0.822 {\scriptsize $\pm$ 0.005} \\
        Corner-based w/ 2D Harris & 0.707 {\scriptsize $\pm$ 0.003} & \textbf{0.886 {\scriptsize $\pm$ 0.008}} & 0.823 {\scriptsize $\pm$ 0.001} \\
        Causal Density-based & \textbf{0.723 {\scriptsize $\pm$ 0.004}} & \underline{0.883 {\scriptsize $\pm$ 0.009}} & 0.789 {\scriptsize $\pm$ 0.011} \\
        Density-based (mean normalized) & --- & --- & 0.821 {\scriptsize $\pm$ 0.008} \\
        \bottomrule
    \end{tabular}
    }
\end{table}

\begin{table}[t]
    \centering
    \caption{Memory usage (\#units) and computational complexity (MACs) for different subsampling methods. Note that the complexity of the corner-based method is  higher than that of the causal density-based method for the same filter size ($\filtersize{=}\harrisfiltersize$).
    }
    \label{tab:supp:comp_complexities}
          \resizebox{\linewidth}{!}{%
    \begin{tabular}{lcc}
\toprule
Subsampling method   & Memory                                                                                & Computational complexity (MACs)                         \\
\midrule
Spatial              & $\mathcal{O}(1)$                                                                                      & $0$                            \\
Temporal              & $\mathcal{O}(1)$                                                                                      & $0$                            \\
Random               & $\mathcal{O}(1)$                                                                                      & $0$                            \\
Event Count          & $\mathcal{O}\left(\frac{\cameraheight\,\camerawidth}{\ratiospatialv\,\ratiospatialh}\right)$ & $N$                            \\
Corner-based         & $\mathcal{O}(\cameraheight\camerawidth)$                                                              & $\cornerbasedOMAC\,N$  \\
Causal density-based & $\mathcal{O}(\cameraheight\camerawidth)$                                                              & $\densitybasedOMAC\,N$\\
\bottomrule
\end{tabular}
    }
    \vspace{-0.4cm}
\end{table}

\section{Discussion and Conclusion}
\label{sec:discussion}

In this paper, we study the trade-offs between hardware-friendly subsampling methods and their impact on object classification accuracy.
To the best of our knowledge, no prior work has systematically compared these methods in terms of task performance. 
Our analysis offers insights into selecting event rate reduction methods, whether implemented in hardware or software.
Among simpler methods with little computation requirements, random subsampling tends to perform better in more scenarios compared to temporal and spatial subsampling. While spatial subsampling can be highly sensitive to offset effects,
more advanced methods, such as density-based and corner-based subsampling, achieve superior classification performance by selectively preserving more informative events from dense regions or corner areas.
We also observe that high variance in event counts across videos can harm the subsampling methods.
To mitigate this effect, adaptive thresholding or an on-off mechanism for dynamically activating the subsampling method could be helpful.

 
\noindent{\textbf{Limitations.}}  
Our analysis is limited to classification tasks using CNNs on three datasets. 
However, the training procedure for different subsampling methods can be applied to other vision tasks and models as a direction for future research.
The proposed density-based method primarily serves as a proof-of-concept to test the hypothesis of density-based subsampling, rather than aiming at computational optimization.
Future exploration can be considered for optimizing the density-based filtering using techniques from \cite{khodamoradi_onon-space_2021,guo_low_2023, liu2015design} to improve efficiency in both memory usage and computational complexity.
{
    \small
    \bibliographystyle{ieeenat_fullname}
    \bibliography{references, main, other_refs}
}

\clearpage
\setcounter{page}{1}
\maketitlesupplementary

\section{Details of Subsampling Parameters}
\label{sec:supp:subsampling_params}

Table~\ref{tab:supp:subsampling_params} presents the specific parameters used for each subsampling method across various subsampling levels.
The parameters are chosen to ensure that the average number of events $\langle\eventtotal\rangle$  remains similar across different subsampling methods at each subsampling level for each dataset.

\begin{table*}[th]
    \centering
    \caption{Parameters of different subsampling methods for various subsampling levels: from 1 (most \#events) to 6 (least \#events). \textbf{mS}: milliseconds.}
    \label{tab:supp:subsampling_params}
         \resizebox{0.95\linewidth}{!}{
         \begin{tabular}{lllcccccc}
\toprule
\multirow{2}{*}{Subsampling methods}  & \multirow{2}{*}{parameters}             & \multirow{2}{*}{dataset} & \multicolumn{6}{c}{subsampling levels (1: most \#events, 6: least \#events)}                          \\ \cline{4-9} 
                                      &                                         &                          & 1             & 2              & 3              & 4               & 5               & 6               \\ \hline\hline
Spatial& ($\ratiospatialh$,$\ratiospatialv$)     & same for all                    & (2,2)         & (4,3)          & (6,6)          & (12,10)         & (15,12)         & (25,16)         \\ \hline
\multirow{2}{*}{Temporal} & $\ratiotemporal$                        & same for all                    & 4             & 12             & 36             & 120             & 180             & 400             \\
                                      & $\temporalwindow$ (mS)                  & same for all                    & 10            & 10             & 10             & 10              & 10              & 10              \\ \hline
Random                    & $\ratiorandom$                          & same for all                    & $\frac{1}{4}$ & $\frac{1}{12}$ & $\frac{1}{36}$ & $\frac{1}{120}$ & $\frac{1}{180}$ & $\frac{1}{400}$ \\ \hline
\multirow{2}{*}{Event Count}          & ($\ratiospatialh$,$\ratiospatialv$)     & same for all                    & (2,2)         & (4,3)          & (6,6)          & (12,10)         & (15,12)         & (25,16)         \\
                                      & $\eventcountthresh$                     & same for all                    & 0.75          & 0.75           & 1.0            & 1.0             & 1.0             & 1.0             \\ \hline
\multirow{7}{*}{Corner-based}         & $\harrisfiltersize$                     & same for all                    & $7\times 7$   & $7\times 7$    & $7\times 7$    & $7\times 7$     & $7\times 7$     & $7\times 7$     \\
                                      & $\cvharrisksize$                        & same for all                    & 3             & 3              & 3              & 3               & 3               & 3               \\
                                      & $\cvharrisblocksize$                    & same for all                    & 2             & 2              & 2              & 2               & 2               & 2               \\
                                      & $\cvharrisk$                            & same for all                    & 0.04          & 0.04           & 0.04           & 0.04            & 0.04            & 0.04            \\ \cline{2-9} 
                                      & \multirow{3}{*}{$\harristhresh$}  & N-Caltech101             & 0.067         & 0.23           & 0.68           & 1.52            & 3.85            & 9.10            \\
                                      &                                         & N-Cars                   & 0.091         & 0.25           & 0.56           & 1.0             & 1.67            & 2.5             \\
                                      &                                         & DVS-Gesture              & 0.077         & 0.17           & 0.5            & 16.7            & 3.33            & 7.70            \\ \hline
\multirow{4}{*}{Causal density-based} & $\temporaldecay$ (mS)                   & same for all                    & 30            & 30             & 30             & 30              & 30              & 30              \\
                                      & $\filtersize$                           & same for all                    & $7\times 7$   & $7\times 7$    & $7\times 7$    & $7\times 7$     & $7\times 7$     & $7\times 7$     \\ \cline{2-9} 
                                      & \multirow{2}{*}{$\threshdensity$} & N-Caltech101 \& N-Cars     & 3.33          & 10.0           & 30.0           & 66.67           & 166.67          & 400.0           \\
                                      &                                         & DVS-Gesture              & 4.63          & 11.63          & 38.56          & 111.11          & 250.0           & 555.56          \\ \bottomrule
\end{tabular}
    }
\end{table*}

\section{Memory Usage and Computational Complexity}
\label{sec:supp:comp_complexities}

In Table~\ref{tab:supp:comp_complexities}, we compare the memory usage and computational complexity  across six different subsampling methods.
We report the total memory units required for an event camera of size $\cameraheight\times\camerawidth$, and computational  complexity in terms of multiply-accumulate operations (MACs)  for a video with $\eventtotal$ number of events.
Spatial, temporal, and random subsampling require only $\mathcal{O}(1)$ memory for  storing a few method-specific parameters and essentially no specific MAC operation.
Event Count method uses memory proportional to the downscaled spatial grid size
$(\frac{\cameraheight}{\ratiospatialv})\times(\frac{\camerawidth}{\ratiospatialh})$, and need $N$ MAC operation for computing the normalized event count per each incoming event.
The corner-based subsampling method adapted from \cite{glover_luvharris_2022} requires  $\mathcal{O}(\cameraheight\camerawidth)$ memory for the event representation frame.
The MAC number contains vertical and horizontal Sobel filtering
$2\,\cvharrisksize^2\,\harrisfiltersize^2\,\eventtotal$, appyling filtering for computing the structural tensor $3\,\cvharrisblocksize^2\,\harrisfiltersize^2\,\eventtotal$, and $10\,\harrisfiltersize^2\,\eventtotal$ for other computations including Harris score calculation.
 The causal density-based subsampling also requires $\mathcal{O}(\cameraheight\camerawidth)$ memory for the method explained in Subsection~\ref{subsec:subsampling-types}. 
The computational complexity is $\densitybasedOMAC\,\eventtotal$ MAC operations.

For an exemplary comparison between the computational operations of the corner-based and causal density-based methods, based on the chosen parameters in Table~\ref{tab:supp:subsampling_params}, the per-event computing cost for the corner-based method is 
$40\,\harrisfiltersize^2$, while for the causal density-based method, it is $4\,\filtersize^2$, where $\harrisfiltersize = \filtersize = 7$.

It is important to note that in Subsection~\ref{subsec:subsampling-types}, our focus was not on optimizing memory usage or computational complexity but rather on analyzing the accuracy performance of density-based methods across different subsampling levels. 
There are existing approaches aimed at developing efficient spatiotemporal filtering methods \cite{khodamoradi_onon-space_2021, guo_low_2023, liu2015design} that can improve both memory efficiency and computational complexity. 
For instance, in \cite{khodamoradi_onon-space_2021}, the authors proposed a spatiotemporal filtering technique that reduces memory usage from $\mathcal{O}(\cameraheight\camerawidth)$ to $\mathcal{O}((\cameraheight\camerawidth)^\frac{1}{2})$.

\section{Visualization of subsampling methods}

\begin{figure*}[tb]
\centering
\resizebox{!}{0.45\textheight}{
\begin{tabular}{@{}l@{}M{50mm}@{}M{50mm}@{}M{50mm}@{}M{50mm}@{}M{50mm}@{}M{50mm}@{}}
& {DVS-Gesture} & {DVS-Gesture} & {N-Caltech101} & {N-Caltech101} & {N-Cars} & {N-Cars} \\
\raisebox{-0.5\height}{\rotatebox{90}{Original}}&
\includegraphics[width=50mm]{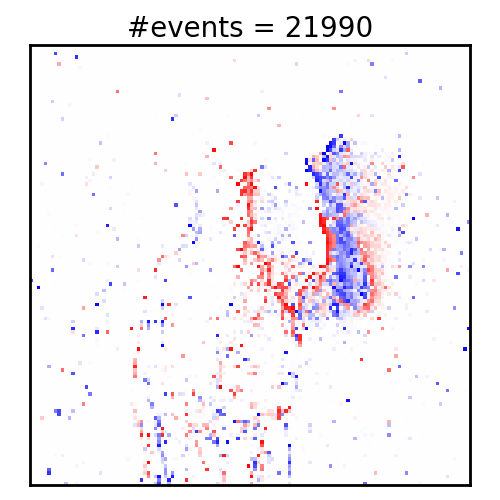} & 
\includegraphics[width=50mm]{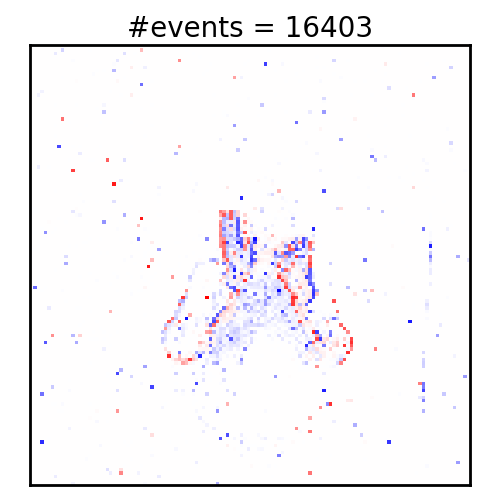} & 
\includegraphics[width=50mm]{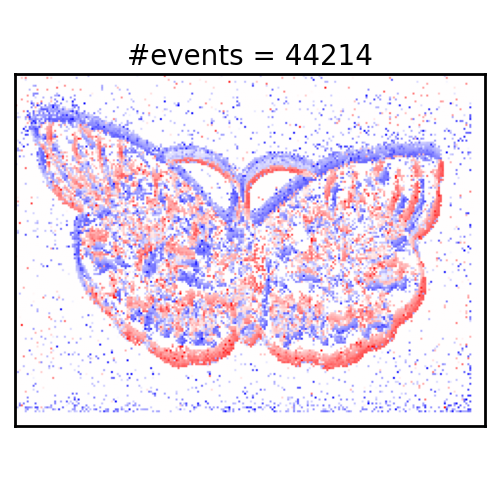} & 
\includegraphics[width=50mm]{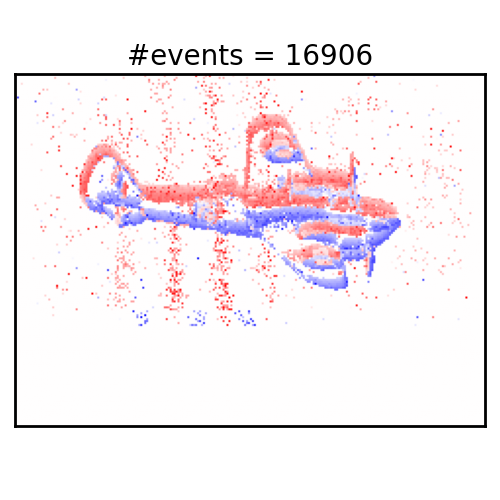} & 
\includegraphics[width=50mm]{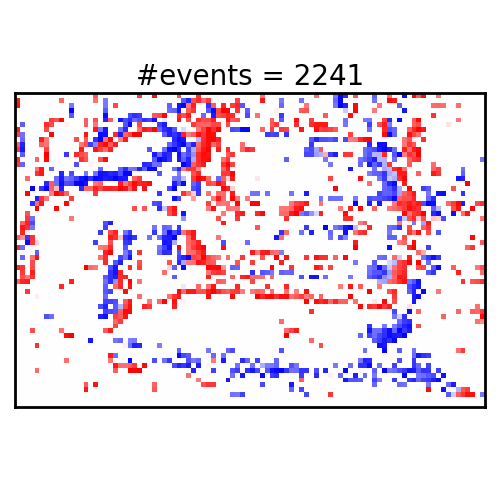} & 
\includegraphics[width=50mm]{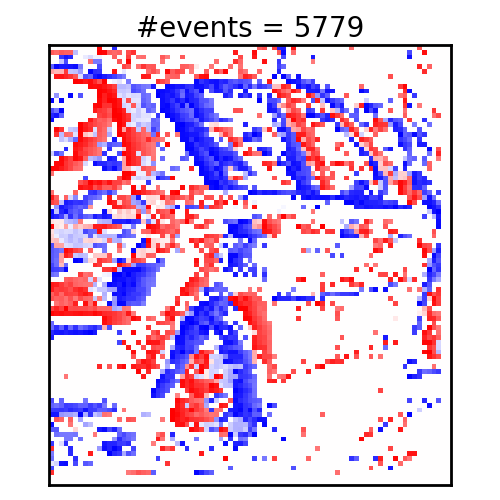} 
\\
\raisebox{-.5\height}{\rotatebox{90}{Spatial}}&
\includegraphics[width=50mm]{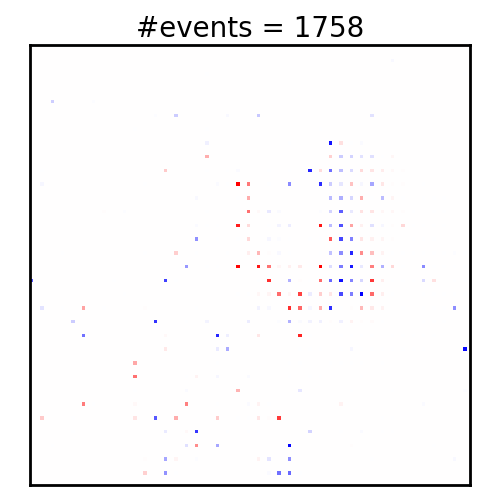} & 
\includegraphics[width=50mm]{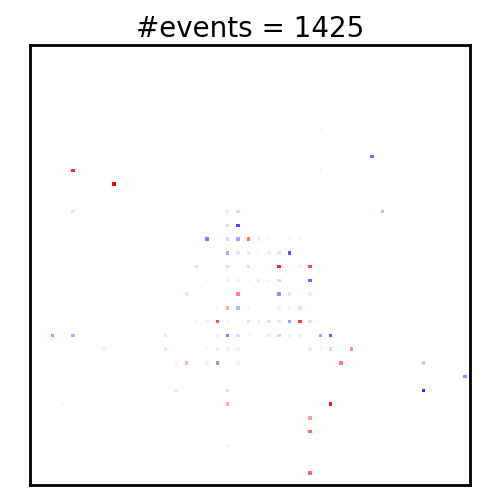} & 
\includegraphics[width=50mm]{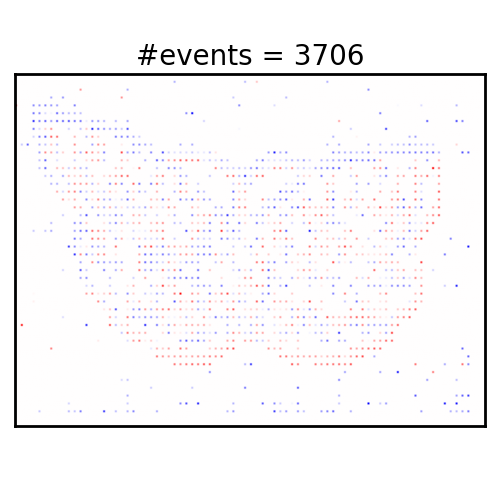} & 
\includegraphics[width=50mm]{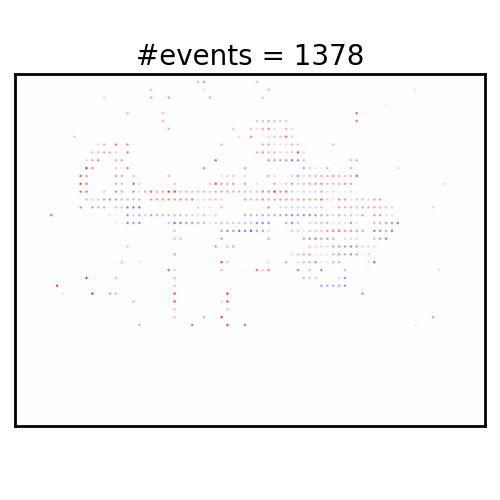} & 
\includegraphics[width=50mm]{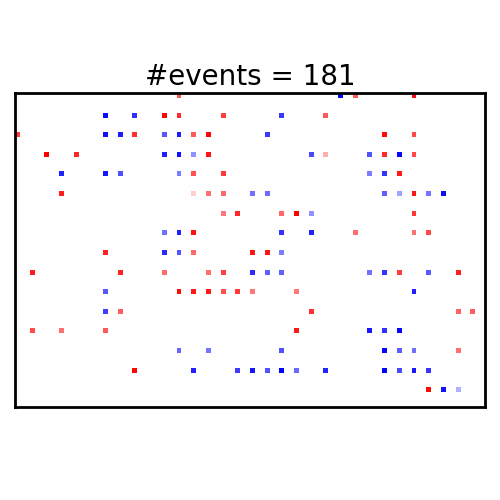} & 
\includegraphics[width=50mm]{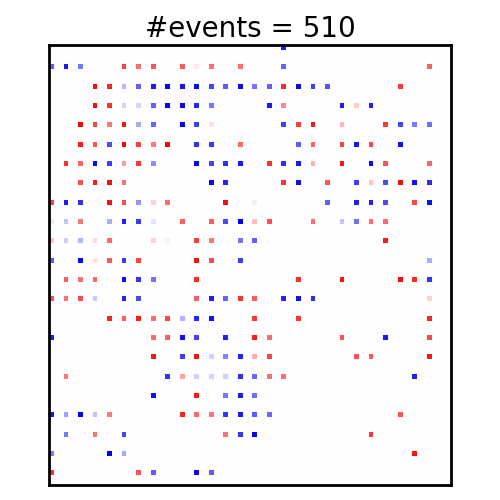} 
\\
\raisebox{-.5\height}{\rotatebox{90}{Temporal}}&
\includegraphics[width=50mm]{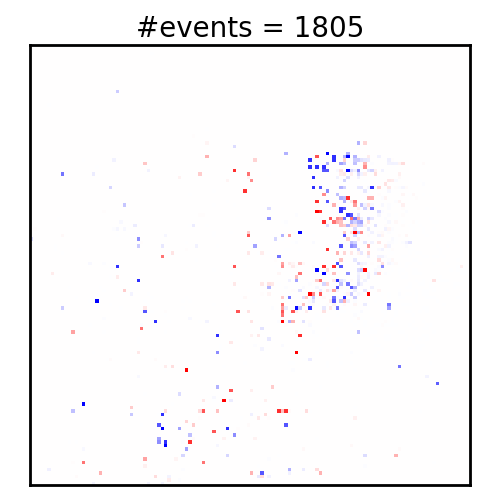} & 
\includegraphics[width=50mm]{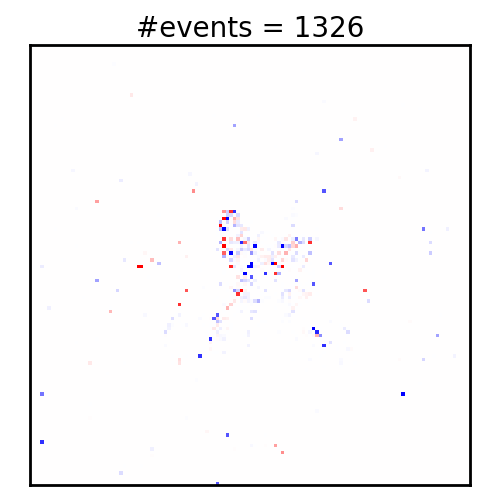} & 
\includegraphics[width=50mm]{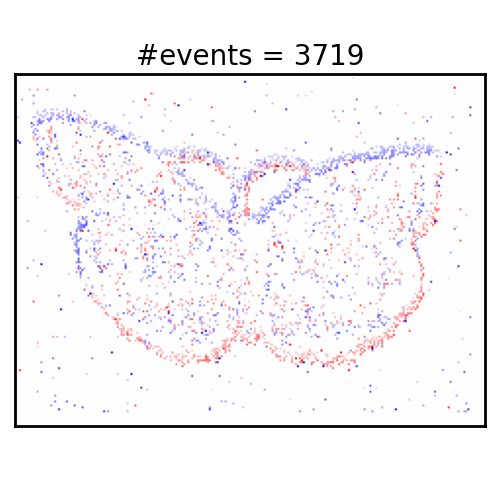} & 
\includegraphics[width=50mm]{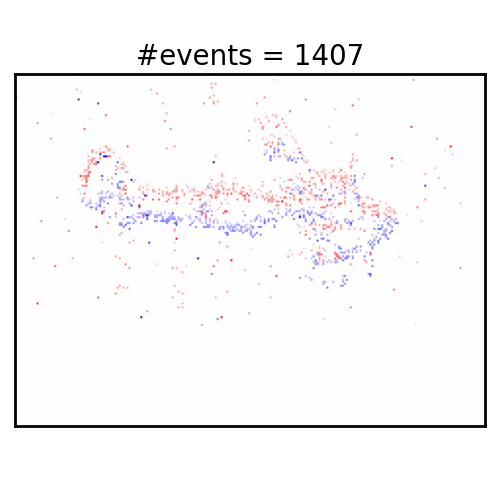} & 
\includegraphics[width=50mm]{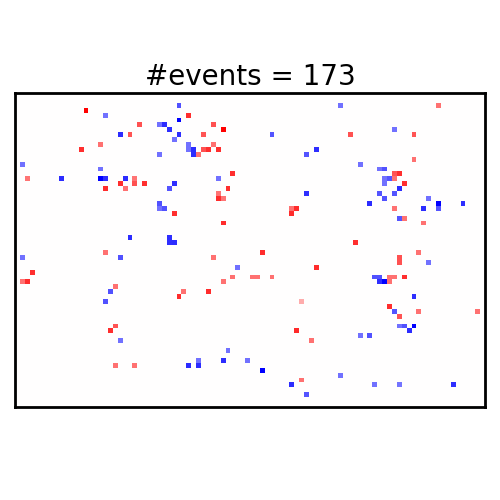} & 
\includegraphics[width=50mm]{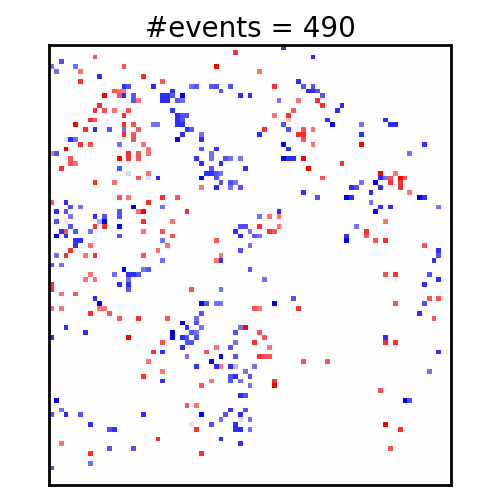} 
\\
\raisebox{-.5\height}{\rotatebox{90}{Random}}&
\includegraphics[width=50mm]{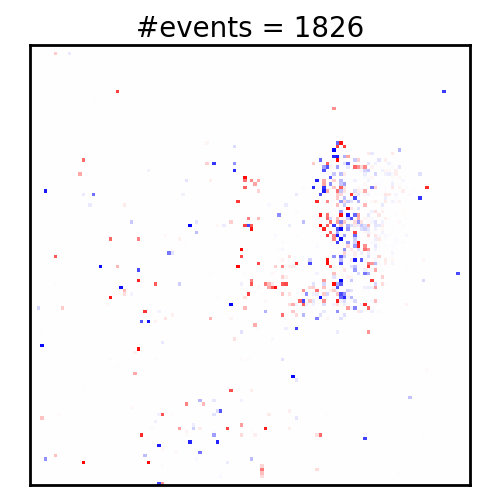} & 
\includegraphics[width=50mm]{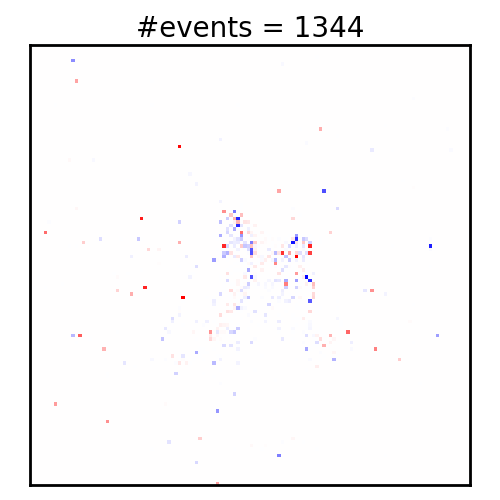} & 
\includegraphics[width=50mm]{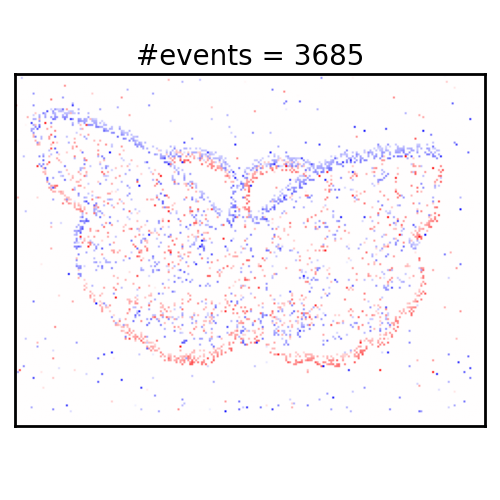} & 
\includegraphics[width=50mm]{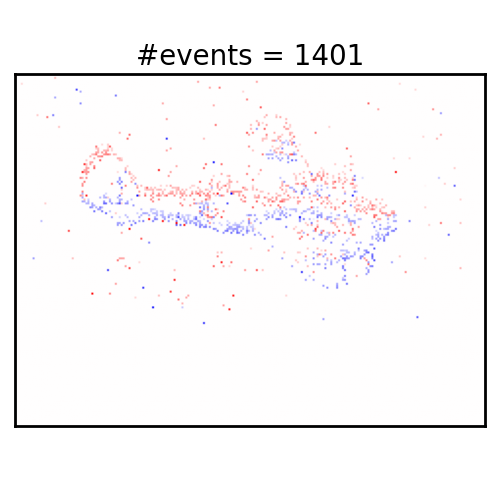} & 
\includegraphics[width=50mm]{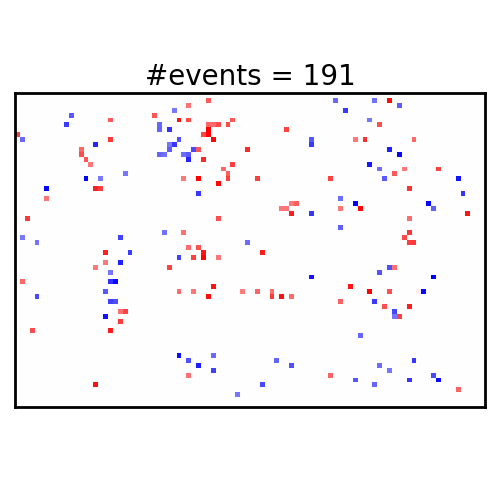} & 
\includegraphics[width=50mm]{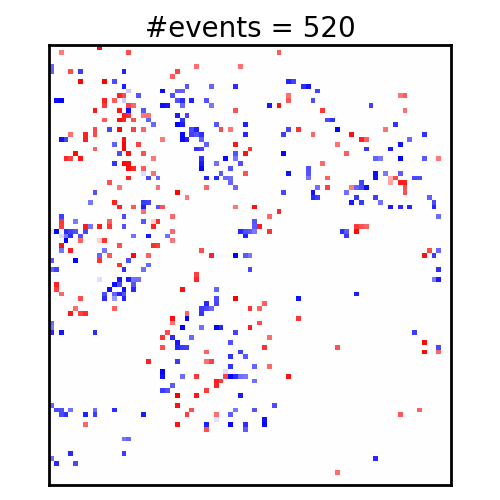} 
\\
\raisebox{-.5\height}{\rotatebox{90}{Event Count}}&
\includegraphics[width=50mm]{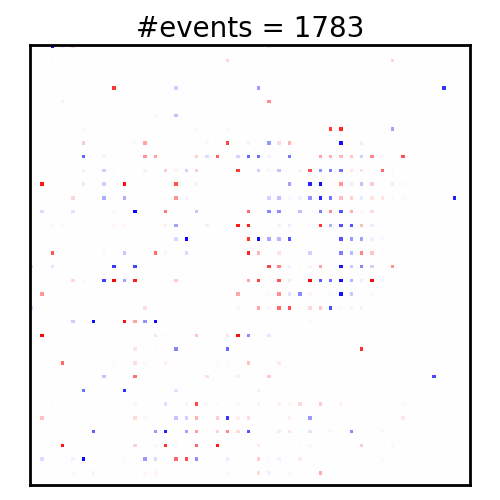} & 
\includegraphics[width=50mm]{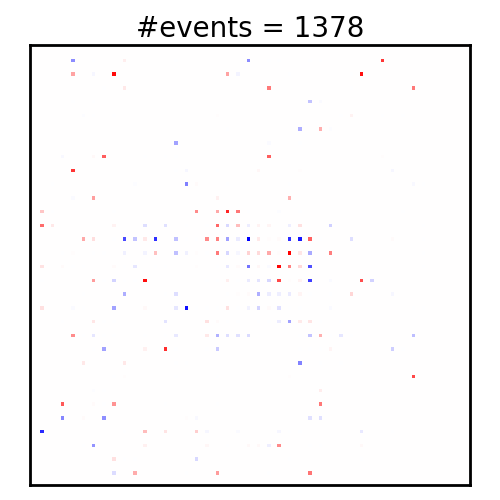} & 
\includegraphics[width=50mm]{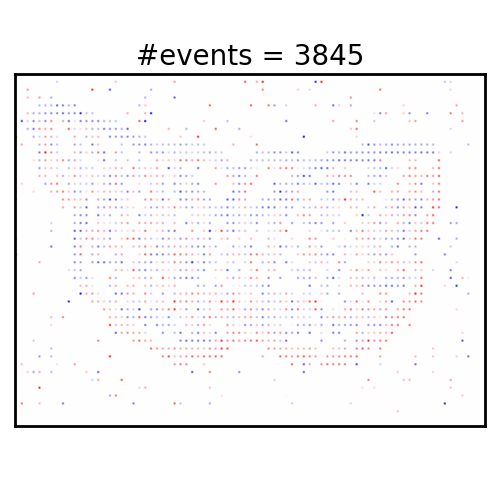} & 
\includegraphics[width=50mm]{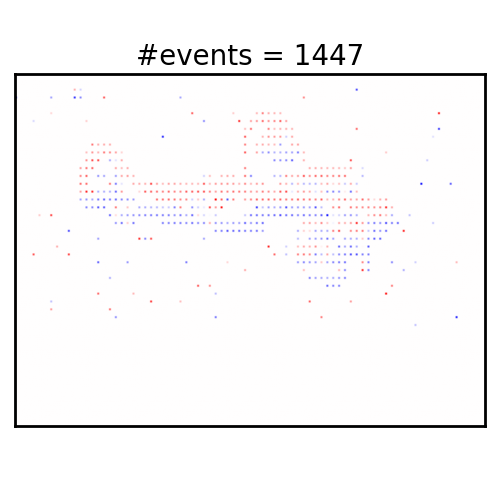} & 
\includegraphics[width=50mm]{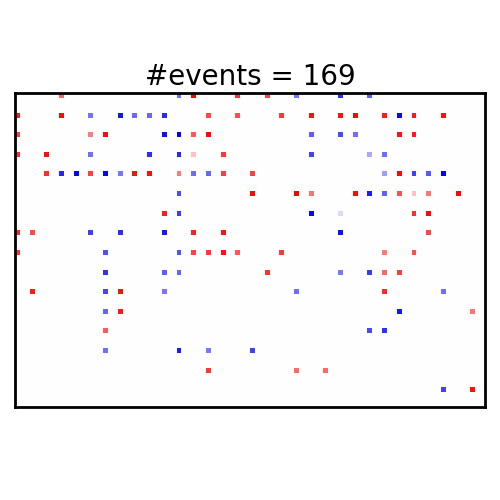} & 
\includegraphics[width=50mm]{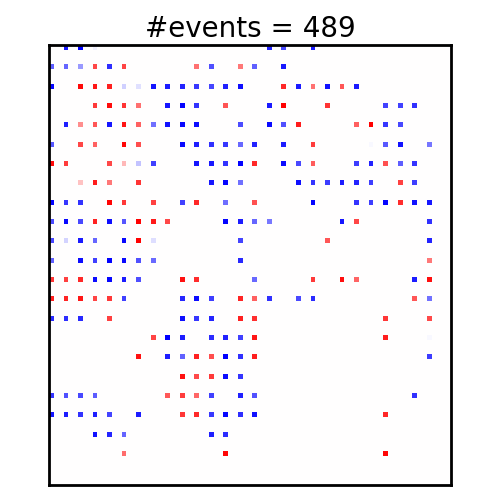} 
\\
\raisebox{-.5\height}{\rotatebox{90}{Causal Density-based}}& 
\includegraphics[width=50mm]{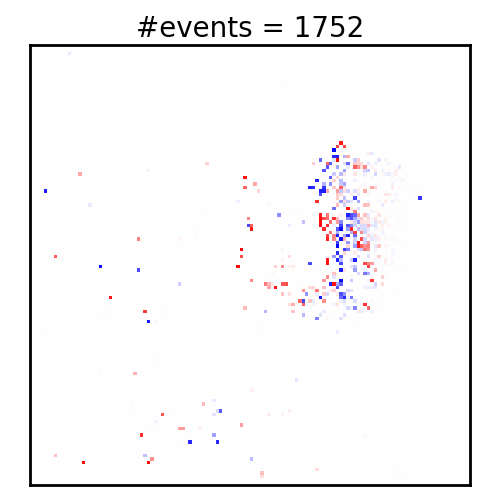} & 
\includegraphics[width=50mm]{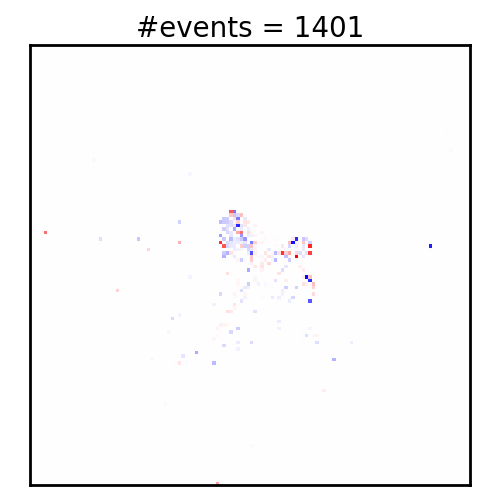} & 
\includegraphics[width=50mm]{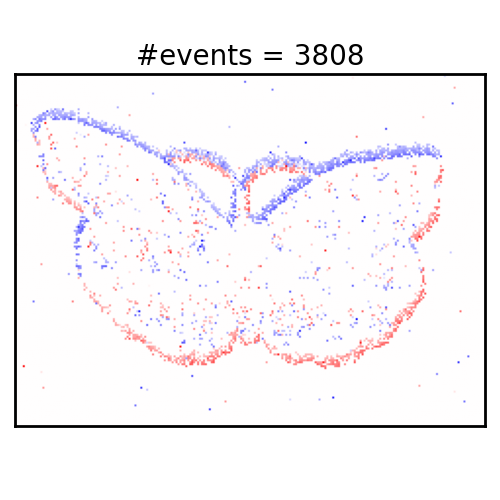} & 
\includegraphics[width=50mm]{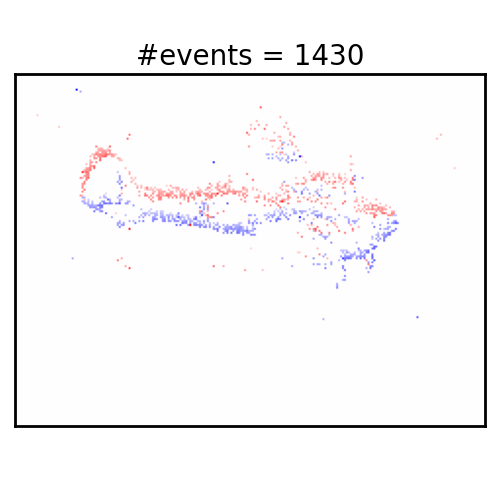} & 
\includegraphics[width=50mm]{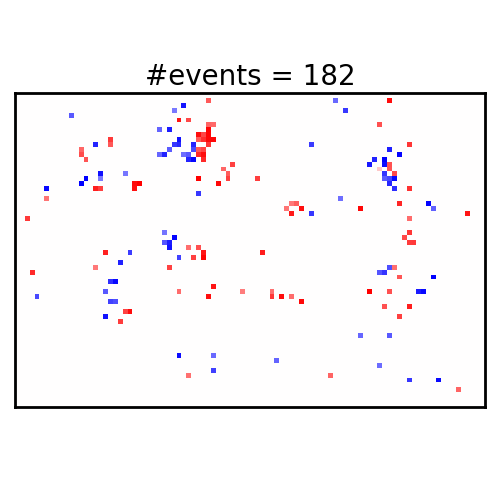} & 
\includegraphics[width=50mm]{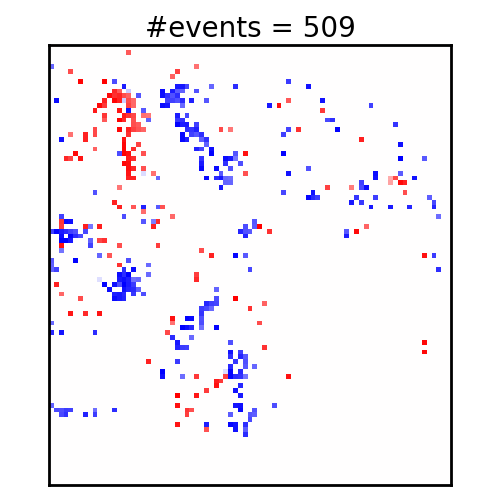} 
\\
\raisebox{-.5\height}{\rotatebox{90}{Corner-based}}& 
\includegraphics[width=50mm]{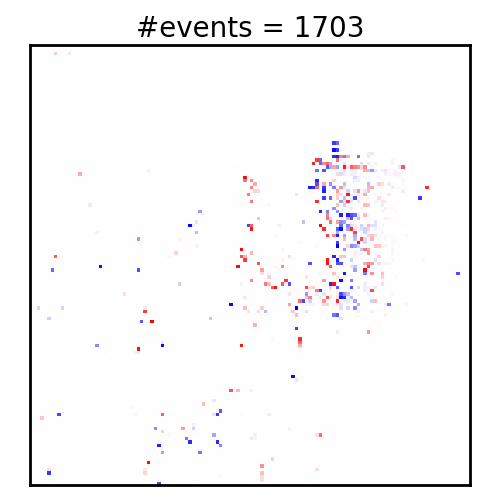} & 
\includegraphics[width=50mm]{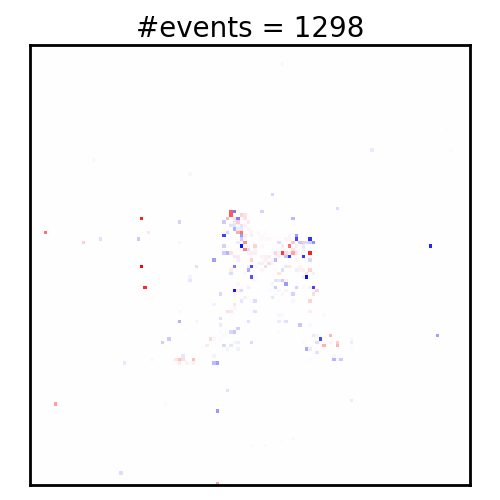} & 
\includegraphics[width=50mm]{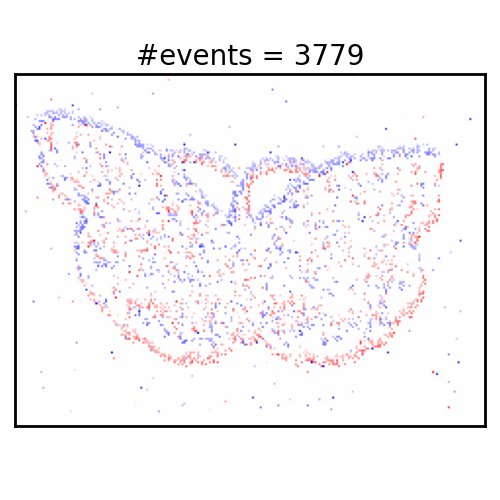} & 
\includegraphics[width=50mm]{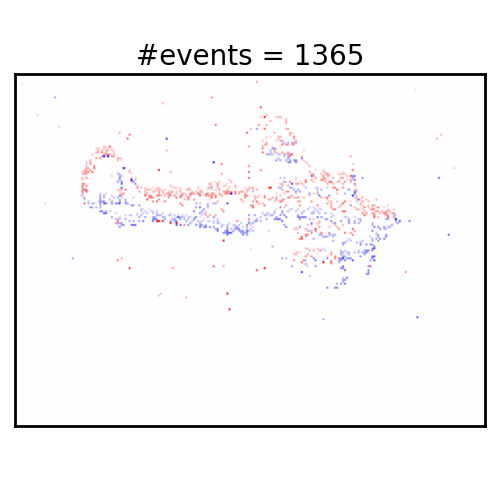} & 
\includegraphics[width=50mm]{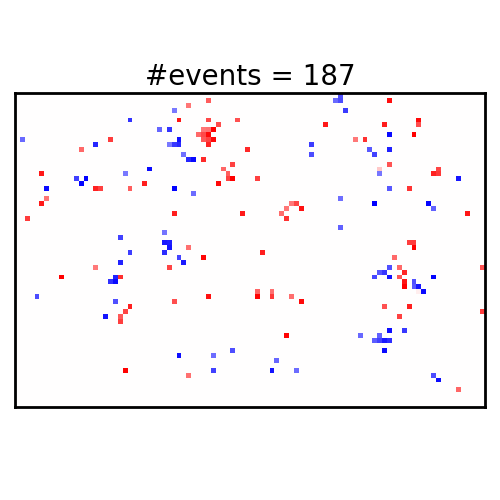} & 
\includegraphics[width=50mm]{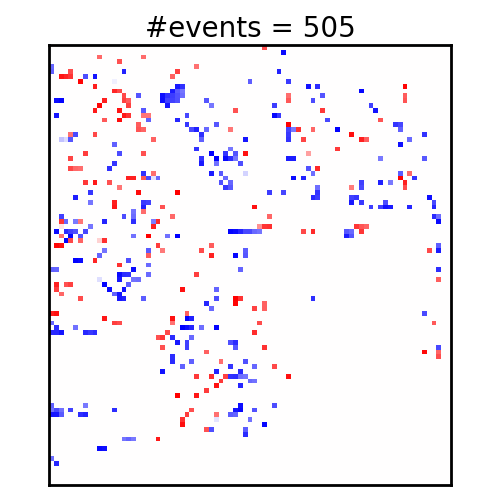} 
\end{tabular}
}
\caption{Visualization of different subsampling methods (starting from the second row). The first row shows the original data. We show for two videos for each dataset. The title of each image is the number of subsampled events. The number of events for different subsampling methods are similar.}
\label{fig:supp:visualization}
\end{figure*}

Figure~\ref{fig:supp:visualization} provides a visualization of the effect of different subsampling methods on event data. In the first row, we show the original event data without any subsampling. Starting from the second row, the figure illustrates the results for two different subsampling levels applied to each dataset. Each image is labeled with the corresponding number of subsampled events, which are consistent across the different subsampling methods.

\end{document}